# Knowledge-Data-Dual-Driven Reinforcement Learning for Autonomous Vehicle Control in Mixed Traffic

Jie Fang, Wei Zheng, Mengyun Xu*, Eui-Jin Kim*

***Abstract*—In mixed traffic, decision-making for autonomous vehicles (AVs) confronts three interrelated challenges. First, physics-based priors incorporated into reinforcement learning (RL) models fail to capture latent interactive vehicle intentions and diverse driver behaviors, limiting the proactive reasoning capabilities. Second, abrupt maneuvers by surrounding vehicles cause non-stationarity, leaving long-tail safety events under-explored. Third, hybrid action spaces destabilize unified RL training due to the different temporal scales of continuous car-following and discrete lane-changing maneuvers. To address these issues, we propose Knowledge-Data Dual-driven Reinforcement Learning (KDDRL). First, a conditional deep generative model synthesizes intention-aware future trajectories, converting passive perception into proactive predictive states. Second, a knowledge-data dual-driven paradigm operates on these predictive states, fusing probabilistic data-driven insights with physical constraints to guide safe exploration through safety-critical scenarios. Third, a coupling module compresses both intention-aware trajectories and physical constraints into compact shared embeddings. This unified representation enables asynchronous multi-timescale optimization of continuous car-following and discrete lane-changing while preserving mutual information. Evaluations on dataset-calibrated simulations demonstrate that KDDRL effectively handles intention uncertainty, accelerates training convergence, and outperforms conventional baseline methods in terms of safety, efficiency, and comfort.**

## I.INTRODUCTION

The rapid advancement of artificial intelligence technologies provides crucial technical support for Autonomous Vehicles (AVs) decision-making systems (Yang et al., 2025). As the core hub connecting environmental perception and vehicle execution, the decision-control system directly governs AV safety, traffic efficiency and driving comfort in complex scenarios (Chen et al., 2022). In mixed traffic environments, an AV interacts with multiple surrounding human-driven vehicles (HDVs) via on-board sensors. By executing cooperative control actions including lateral lane-changing and longitudinal car-following, the controlled ego AV optimizes its own driving behavior while adapting to the stochastic motions of neighboring vehicles, as shown in **Fig. 1**. Existing research indicates that lane-changing and car-following behaviors exhibit strong dynamic coupling, yet they involve inherently different action characteristics, discrete versus continuous, that are difficult to capture using either identical or completely isolated control models. Integrating both behaviors into a unified framework with strategies tailored to their distinct characteristics represents a key pathway to enhancing AV decision-making capabilities (Chen et al., 2024; Ali et al., 2025).

Current control methods for individual AVs in mixed traffic scenarios are primarily categorized into rule-based and data-driven models. Rule-based approaches rely on manually preset parameters and traffic rules, suffering from limited adaptability and poor generalization in complex dynamic environments (Jin et al., 2021;Wu et al., 2013). In contrast, data-driven approaches, particularly reinforcement learning (RL), optimize driving strategies through continuous agent-environment interaction (Archives et al., 2022; Li et al., 2022). Recently, integrating the interpretability of rule-based models into RL, termed physics-informed RL, has demonstrated superior control performance and has become a mainstream trend in vehicle control research.

Although RL has demonstrated advantages in autonomous driving decision-making, existing physics-informed methods still suffer from three key limitations. (1) **Deterministic physics models incorporated into RL overlook latent interactive vehicle intentions and diverse driver behaviors** (Lin et al., 2024; Özsüer et al., 2023)**.** Most existing studies incorporate trajectory prediction based on classical physical models, such as kinematic equations or minimizing overall braking induced by lane change (MOBIL) (Shu et al., 2022; Mo et al., 2022), producing only deterministic, physically feasible trajectories. These methods fail to capture the behavioral divergence of human drivers arising from different intentions under identical physical states. For instance, given the same traffic gap, some drivers choose to yield while others execute aggressive cut-ins. Consequently, RL agents lose the ability to proactively reason about the future evolution of surrounding vehicles, hindering anticipatory decisions in highly interactive scenarios. (2) **RL-based decision making suffer from non-stationarity during safety-critical interactions, as surrounding vehicles may act abruptly, causing distributional shifts and leaving long-tail safety events under-explored** (Zhao et al., 2024)**.** Safety-critical scenarios, such as emergency braking and sudden cut-ins, trigger fundamentally different responses: some drivers yield aggressively while others execute evasive steering. These abrupt behavioral divergences during safety-critical scenarios

[1]This work was supported by the National Research Foundation of Korea (NRF) grant funded by the Korea government (MSIT) (No.RS-2024-00337956). Jie Fang was supported by National Natural Science Foundation of China under Grants 71901070. Mengyun Xu was supported by Natural Science Foundation of Fujian Province under Grants 83026085. (Corresponding authors: Mengyun Xu and Eui-Jin Kim.)

Jie Fang and Wei Zheng is department of Civil Engineering, Fuzhou University, Fuzhou, 350108, China (e-mail: fangjie@fzu.edu.cn ; 240527278@fzu.edu.cn ).

Mengyun Xu is department of Civil Engineering, Fuzhou University, Fuzhou, 350108, China and department of Civil and Environmental Engineering, National University of Singapore, 119077, Singapore(e-mail: xumengyun@fzu.edu.cn ).

Eui-Jin Kim is department of Transportation Systems Engineering, Ajou University, Suwon 16499, Republic of Korea(e-mail: euijin@ajou.ac.kr ).

violate the stationarity required for stable policy optimization in RL. Without explicit mechanisms to explore these abrupt behavioral variations, the agent lacks sufficient exposure to rare but safety-critical events. (3) **Continuous acceleration for car-following and discrete lane-changing operate at fundamentally different temporal resolutions, creating hybrid action spaces that destabilize unified RL training when forced into synchronous updates, while naive decoupling fails to capture interdependent information.** Existing unified frameworks that apply identical RL methods to both control behaviors often face training instability. Conversely, simple decoupling into independent modules further isolates decision context, compromising coordinated control (Wei et al., 2024; Lin et al., 2024).

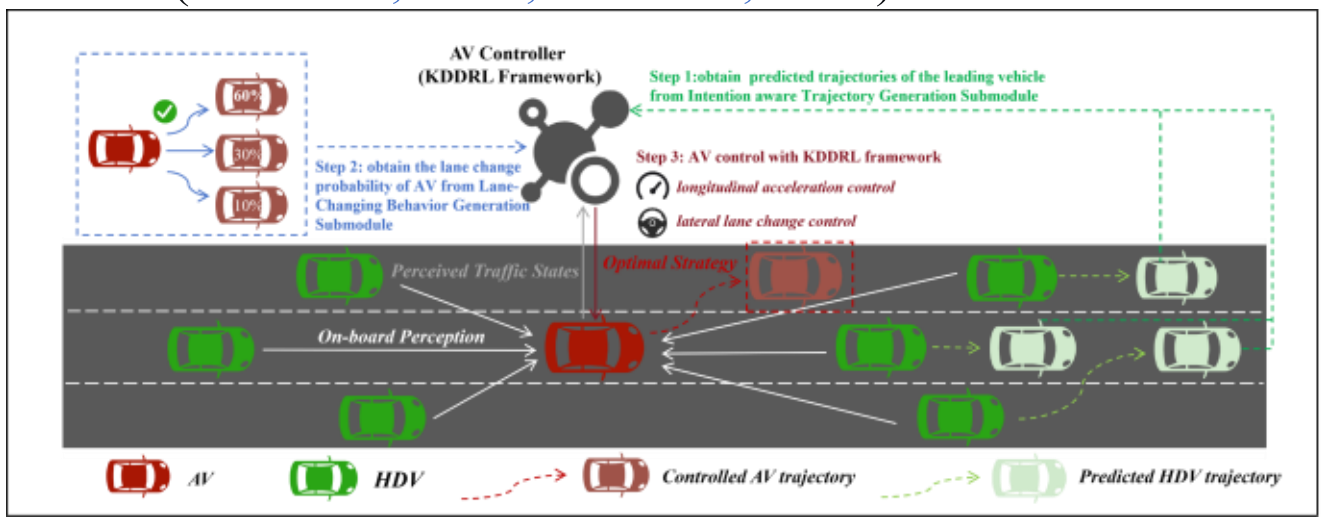

**Fig.1.** Schematic framework of the proposed KDDRL for single-ego AV control in mixed traffic scenario

To address these gaps, this paper proposes a Knowledge-Data Dual-driven Reinforcement Learning (KDDRL) framework for the control of a single ego AV in mixed traffic (**Fig.1**). The core design philosophy shifts from passive perception-reaction to proactive prediction-reasoning through three synergistic transformations. First, instead of feeding raw observations directly into RL (Cui et al., 2024; Ni et al., 2024), we introduce intention-aware predictive states that explicitly model surrounding vehicles' behavioral uncertainty, thereby enabling anticipatory decision-making. Second, rather than treating physical knowledge as hard constraints that may restrict learning (Huang et al., 2025), we embed interpretable physical boundaries as inductive biases within data-driven learning, so that empirical driving logic and physical plausibility jointly guide exploration without sacrificing adaptability to diverse human behaviors. Third, instead of forcing heterogeneous actions into synchronous optimization or decoupling them into isolated modules (Peng et al., 2022), we establish a unified predictive-knowledge state basis that preserves interdependent information across decision types while allowing each to operate at its natural temporal solution. This ensures that lane-changing and car-following remain coordinated despite their asynchronous optimization at different time scales. The influence of aforementioned three core design concepts on AV control is illustrated in **Fig. 1**. The specific contributions of this paper are summarized as follows:

- **An improved intention-aware trajectory generation model:** Unlike existing methods that rely on deterministic physics models, this study proposes a conditional generative adversarial network (cGAN) that encodes interactive intentions and synthesizes multiple physically plausible trajectories. These predictive states convert passive perception into intention-aware representations, providing a rich state basis for proactive decision-making.
- **Training on predictive states with embedded data and physics priors:** To address non-stationarity and under-explored long-tail events, this study designs a knowledge-data dual-driven paradigm that extracts data-driven insights from probabilistic distributions while employing interpretable physical boundaries as safety-efficiency inductive biases. This coupling stabilizes policy optimization across both nominal and safety-critical non-stationary regimes while improving exploration efficiency in rare safety-critical events.
- **Shared predictive-knowledge state basis for decoupled heterogeneous control:** To resolve the instability of hybrid discrete-continuous action spaces, a coupling module compresses intention-aware trajectories and physics-based constraints into compact shared embeddings, establishing a unified predictive-knowledge state basis. These embeddings facilitate asynchronous optimization for continuous car-following and discrete lane-changing at different time scales. This temporal decoupling stabilizes training, while the shared representation ensures that longitudinal and lateral maneuvers remain mutually informed.

## II. RELATED WORKS AND PRELIMINARIES

Existing vehicle control methods, encompassing car-following and lane-changing behaviors, can be broadly classified into three categories: rule-based model-driven approaches, reinforcement learning (RL)-based interactive learning methods, and physics-informed RL (PIRL) models that integrate the strengths of both.

### *A. Rule-based Vehicle Control Methods*

Rule-based model-driven methods rely on manually preset parameters and physical modeling to describe vehicle behaviors. The classic Intelligent Driver Model (IDM) characterizes car-following behavior through dynamic equations and can be calibrated to adapt to different driving styles (Treiber et al., 2000); however, its deterministic framework struggles to cope with stochastic fluctuations and driving heterogeneity. The MOBIL model performs lane-changing decisions through safety distance verification and utility functions, offering good interpretability and scalability (Kesting et al., 2007). While these methods are computationally efficient and provide controllable safety, they are inherently limited in real-world scenarios characterized by high uncertainty, multimodal distributions, and driving heterogeneity. More fundamentally, deterministic physics models fail to capture the behavioral divergence of human drivers, where different intentions can lead to vastly different responses under identical physical states. Consequently, rule-based methods deprive downstream decision-making systems of the ability to proactively reason about surrounding vehicles' future evolution.

### *B. RL-Based Vehicle Control Methods*

In contrast to the above rule-driven methods, RL optimizes policies through continuous interaction between agents and the environment. RL requires no preset rules and can explore optimal policies in a data-driven manner, exhibiting strong robustness in complex scenarios (Huang et al., 2025; Liu and Yu, 2025). Existing RL research has often modeled lane-changing and car-following in a unified, synchronous manner. For instance, Cui et al. (2024) proposed an integrated lateral and longitudinal decision model that uses a deep Q-network to handle both lane-changing commands and car-following acceleration simultaneously. Similarly, Lin et al. (2024) utilized a parameterized soft Actor-Critic (PASAC) method to output discrete lane-changing decisions and continuous longitudinal acceleration within a single framework, aiming for a unified optimization of hybrid actions. However, continuous longitudinal car-following and lateral discrete lane-changing operate at fundamentally different temporal resolutions; the former requires high-frequency continuous adjustments, while the latter involves intermittent, discrete maneuvers. When forced into synchronous updates, such hybrid action spaces often destabilize unified RL optimization. Moreover, naive decoupling of these two behaviors fails to capture interdependent information, leading to sub-optimal coordination.

To address this, researchers have adopted hierarchical control architecture. Peng et al. (2022) developed a hierarchical framework with separate agents for discrete lane-changing and continuous car-following decisions, yet both operate at the same control interval. Recognizing the temporal asymmetry between these two behaviors, Ni et al. (2024) adopted a heterogeneous- frequency framework. While this partially addresses the temporal mismatch, the cooperative mechanism between the two agents remains loosely coupled, and the shared state representation does not systematically encode the mutual dependence between lane-changing feasibility and car-following safety margins. Most importantly, these methods typically feed raw current-step observations into the RL agents. Without explicit modeling of future trajectory uncertainty and the interaction intentions of surrounding vehicles, the agent lacks exposure to rare but hazardous situations during training, causing severe distributional shifts and impeding stable learning. This motivates the integration of physics-informed principles into RL-based vehicle control.

### C. Physics-informed RL Control Methods

PIRL has emerged to improve sample efficiency and safety by encoding physical constraints (e.g., vehicle dynamics, traffic rules, safety boundaries) into the RL framework. For example, the PE-RLHF framework (Huang et al., 2025) integrates human feedback and physical knowledge to direct agent exploration toward safer regions, thereby also enhancing its adaptive generalizability across varying scenarios. Similarly, the PIPF-TD3 framework (Pan et al., 2026) combines artificial potential fields with the TD3 algorithm to provide explicit safety guarantees. Peng et al. (2022) employed MOBIL-derived safety-efficiency criteria to improve the safety of D3QN-based lane-changing control.

However, existing PIRL approaches for AV control primarily focus on incorporating deterministic physical knowledge, while overlooking the interactive vehicle intentions and diverse driver behaviors of surrounding vehicles. In mixed traffic flows, future vehicle trajectories are inherently probabilistic and intention-dependent; a vehicle may exhibit diverse behavioral modes under different traffic contexts (Xu et al., 2025). Instead of relying on pure physics-based models, which fail to capture these multimodal distributions, or pure data-driven methods lacking explicit safety guarantees, effective AV control requires a synergetic approach. Combining data-driven state prediction with intention-aware trajectory generation is a critical step toward bridging this gap.

Data-driven methods, particularly Generative Adversarial Networks (GAN), offer unique advantages in characterizing multimodal behaviors due to their ability to fit complex data distributions. GANs learn the latent distribution of real data and generates highly realistic new samples through adversarial training between a generator and a discriminator (Goodfellow et al., 2014; Mirza and Osindero,2014). In vehicle trajectory modeling, Shi et al. (2025) employed an encoder-decoder LSTM-based generator and an LSTM-MLP discriminator to produce realistic trajectories whose distribution closely resembles that of the training data. Xu et al. (2025) proposed combining a conditional deep generative model with a physics-based vehicle following model to reconstruct complete vehicle trajectories with diverse driver behaviors. The introduction of data-driven GAN-based methods enables learning multimodal behavioral distributions from real driving data and generating diverse future trajectories that conform to authentic driving patterns.

Nevertheless, in existing deep generative models, the coupling between interactive intention conditions and physics plausibility constraints remains relatively loose, limiting the applicability of generated trajectories as reliable predictive states for downstream decision-making. To bridge this gap, this study adopts a conditional generative adversarial network to generate future trajectories of the leading vehicle based on the interaction intentions between the ego vehicle and surrounding vehicles, thereby providing forward-looking predictive state representations for the RL agent. Building on these predictive states, a knowledge-data dual-driven paradigm is established that fuses data-driven probabilistic distributions with interpretable physical boundaries, operating on predictive states rather than raw observations to enhance policy optimization and improve exploration efficiency in rare safety-critical events.

### D. Summary and Research Positioning

The foregoing review identifies three interrelated gaps. Firstly, deterministic physics models (IDM, MOBIL) and existing data-driven methods fail to explicitly encode interactive vehicle intentions across lanes. They produce either single feasible trajectories or statistically realistic but intention-agnostic samples, depriving RL agents of structured foresight for proactive reasoning. Secondly, RL methods relying on raw perceptual observations suffer from non-stationarity and distribution shift as surrounding vehicles maneuver abruptly. Existing PIRL approaches embed physical knowledge primarily at the level of raw observations or reward functions, without systematically combining data-driven probabilistic predictions with interpretable physical boundaries on a unified predictive state basis. Consequently, long-tail safety events remain under-explored and convergence is inefficient. Thirdly, hybrid discrete-continuous actions for lane-changing and car-following destabilize unified RL when forced into synchronous updates. While hierarchical decoupling partially mitigates temporal mismatch, it does so at the cost of isolating interdependent decision information, and the shared state representation lacks a systematic mechanism to encode the mutual dependence between the two behaviors.

These three limitations are structurally coupled. Intention-unaware prediction produces inaccurate future estimates, which exacerbates observation non-stationarity during training; high-dimensional predictive states combined with heterogeneous action spaces further amplify control instability. Therefore, a piecemeal solution at any single layer is insufficient. To this end, this study proposes the KDDRL framework. It incorporates an intention-aware trajectory generative module to provide forward-looking predictive states, establishes a knowledge-data dual-driven paradigm operating on these states to stabilize optimization and improve exploration in rare events, and devises a unified predictive-knowledge state basis with asynchronous multi-timescale optimization to handle hybrid discrete-continuous actions while preserving interdependent decision context.

*E. Preliminaries of Reinforcement Learning*

To facilitate the subsequent exposition of methods, we first introduce the basic framework of RL. The fundamental framework of RL is the Markov Decision Process (MDP), which can be represented as a tuple $(S, A, R, P, \gamma)$, where $S$ is the state space, $A$ is the action space, $R: S \times A \to \mathbb{R}$ is the reward function, $P: S \times A \times S \to [0,1]$ is the state transition probability distribution, and $\gamma \in [0,1]$ is the discount factor. Model-free RL does not require obtaining the transition model $P$ in advance; instead, it allows the agent to interact with the environment and continuously adjust its policy to achieve higher rewards. This study adopts two RL models, D3QN and TD3, to handle the discrete-continuous hybrid action space. The former generates discrete lane-change commands (right lane change, lane keeping, and left lane change), leveraging value decomposition to assess the relative utility of lateral maneuvers, while the latter selects continuous longitudinal acceleration strategies to ensure smooth speed tracking. As core components of the MDP, the state space, action space, and reward function of the two models will be detailed in Section 3.

***(1) D3QN Model***

Deep Q-Network (DQN) has advantages in handling discrete action spaces. However, the traditional DQN algorithm suffers from Q-value overestimation, leading to unstable policy updates. Van Hasselt proposed the Double Deep Q-Network (DDQN), which effectively alleviates the overestimation problem by decoupling action selection from action evaluation (Van et al., 2016). Wang further proposed the Dueling Deep Q-Network, which decomposes the Q-value function into the sum of a state-value function and an advantage function (Wang et al., 2016). D3QN combines double Q-learning with the dueling architecture, employing primary and target network separation alongside a dueling network decomposition, thereby achieving faster training convergence speed and higher performance stability.

D3QN introduces the state-value function $V^\pi(s)$ and the advantage function $A^\pi(s,a)$ from Dueling DQN, expressed as:

$$V^\pi(s) = \mathbb{E}_{a\sim\pi(s)}[Q^\pi(s,a)] \tag{1}$$

$$A^\pi(s,a) = Q^\pi(s,a) - V^\pi(s) \tag{2}$$

The state-value function $V^\pi(s)$ represents the expected cumulative return from state $s$ under policy $\pi$, evaluating the overall quality of the state. The advantage function $A^\pi(s,a)$ measures the relative importance of executing action $a$ compared to the average return at state $s$. To approximate the true action-value function $Q^\pi(s,a)$, D3QN parameterizes the Q-function using a deep neural network $Q(s,a;\theta,\alpha,\beta)$. To ensure structural identifiability of $V$ and $A$ during training, the dueling architecture decompose $Q(s,a;\theta,\alpha,\beta)$ as:

$$Q(s,a;\theta,\alpha,\beta) = V(s;\theta,\beta) + \left(A(s,a;\theta,\alpha) - \frac{1}{|\mathcal{A}|}\sum_{a'} A(s,a';\theta,\alpha)\right) \tag{3}$$

where $\theta$ denotes the shared network parameters, $\alpha$ is the parameter vector specific to the advantage function, $\beta$ is the parameter for the state-value function, and $|\mathcal{A}|$ is the number of actions. Subtracting the mean average $\frac{1}{|\mathcal{A}|}\sum_{a'} A(s,a';\theta,\alpha)$ forces the advantage function to have a zero mean for chosen actions, eliminating the arbitrary translation degrees of freedom and stabilizing optimization (Wang et al., 2016).

D3QN selects actions using primary network parameters $\theta$ and evaluates their values using target network parameters $\theta'$. The loss function over a mini-batch of size $N$ sampled from replay buffer is defined as:

$$L_{D3QN} = \frac{1}{N}\sum_{i=1}^{N}(r_i + \gamma Q(s_i', argmax_{a'} Q(s_i', a'; \theta, \alpha, \beta); \theta', \alpha', \beta') - Q(s_i, a_i; \theta, \alpha, \beta))^2 \tag{4}$$

The target network parameter $\theta'$ are updated softly using the update rate $\tau$:

$$\theta' \leftarrow (1-\tau)\theta' + \tau\theta \tag{5}$$

The training process of D3QN follows the traditional deep Q-learning paradigm. The agent explores the environment using an $\varepsilon$-greedy policy, stores experience tuples in an experience replay buffer, randomly samples mini-batches to minimize the loss function, and updates the network parameters accordingly.

***(2) TD3 Model***

For the continuous decision problem such as car-following control, policy gradient methods provide smoother policy updates than value-function-based methods. Deep Deterministic Policy Gradient (DDPG) learns a deterministic policy actor $a = \mu(s;\phi)$ and a critic value function through the Actor-Critic framework. However, DDPG suffers from Q-value overestimation and insufficient exploration in continuous action spaces (Timothy et al., 2019; Li et al., 2025). Fujimoto proposed Twin Delayed Deep Deterministic Policy Gradient (TD3), which improves continuous control stability using twin Critic networks, target policy smoothing regularization, and delayed Actor updates (Fujimoto et al., 2018).

The TD3 model consists of one actor network $\mu(s;\phi)$, one target actor network $\mu(s';\phi')$, two critic networks $Q_k(s,a;\varphi_k)$(for $k \in \{1,2\}$), and two target critic networks $Q_k(s',\tilde{a};\varphi_k')$. Target policy smoothing introduces clipped normal noise (its value is clipped to the range $[-c,c]$) to the target action $\tilde{a}$ to prevent value overestimation on narrow peaks. The smoothed target action is computed as:

$$\tilde{a} = \mu(s';\phi') + \epsilon, \epsilon \sim clip(\mathcal{N}(0,\sigma), -c, c) \tag{6}$$

The Critic network parameters $\varphi_k$ are updated by minimizing the mean squared error loss over a mini-batch size $N$ sampled from replay buffer:

$$L_{TD3} = \frac{1}{N}\sum_{i=1}^{N}(r_i + \gamma min_{k=1,2} Q_k(s_i', \tilde{a}; \varphi_k') - Q_k(s_i, a; \varphi_k))^2 \quad (7)$$

The Actor network parameters $\phi$ are updated using the deterministic policy gradient evaluated on the first Critic network $Q_1$:

$$\nabla_\phi J(\phi)_{TD3} = \frac{1}{N}\sum_{i=1}^{N} \nabla_a Q_1(s_i, a; \varphi_1)|_{a=\mu(s_i;\phi)} \nabla_\phi \mu(s_i; \phi) \quad (8)$$

To reduce variance and accumulated error, the Actor network and target networks are updated less frequently than the critic networks (i.e., delayed policy updates). Soft updates for target parameters are expressed as:

$$\varphi_k' \leftarrow (1-\tau)\varphi_k' + \tau\varphi_k, k \in \{1,2\} \quad (9)$$

$$\phi' \leftarrow (1-\tau)\phi' + \tau\phi \quad (10)$$

## III. METHODOLOGY

### *A. KDDRL Framework*

This study proposes the KDDRL framework for the decision-making of a single ego AV exhibiting continuous car-following and discrete lane-changing behaviors in multi-lane mixed traffic scenarios. The overall architecture illustrated in **Fig. 2**, comprises a knowledge-data dual-driven module and a vehicle control module.

Within the knowledge-data dual-driven module, we design an intention-aware trajectory generation submodule to generate physically plausible future trajectories of surrounding leading vehicles with interactive intentions. Concurrently, the lane-changing behavior generation submodule, building upon the generated trajectories is designed to learn human lane-changing behavior logic and produce lane-changing probabilities (i.e., probabilistic lane-changing priors). In the vehicle control module, the generated intention aware trajectories and lane-changing probabilities are integrated with raw observations to form a unified predictive-knowledge state representation, which is fed into the RL agent. The control module employs differentiated RL strategies for heterogeneous actions: a high-frequency TD3 policy (operating at 0.1s intervals) optimizes continuous car-following acceleration, while a low-frequency D3QN policy (operating at 2s intervals) governs discrete lane-changing decisions, achieving asynchronous multi-timescale coordination under the unified state basis. The control actions are executed in the mixed traffic environment, and the updated observations are fed back to close the control loop.

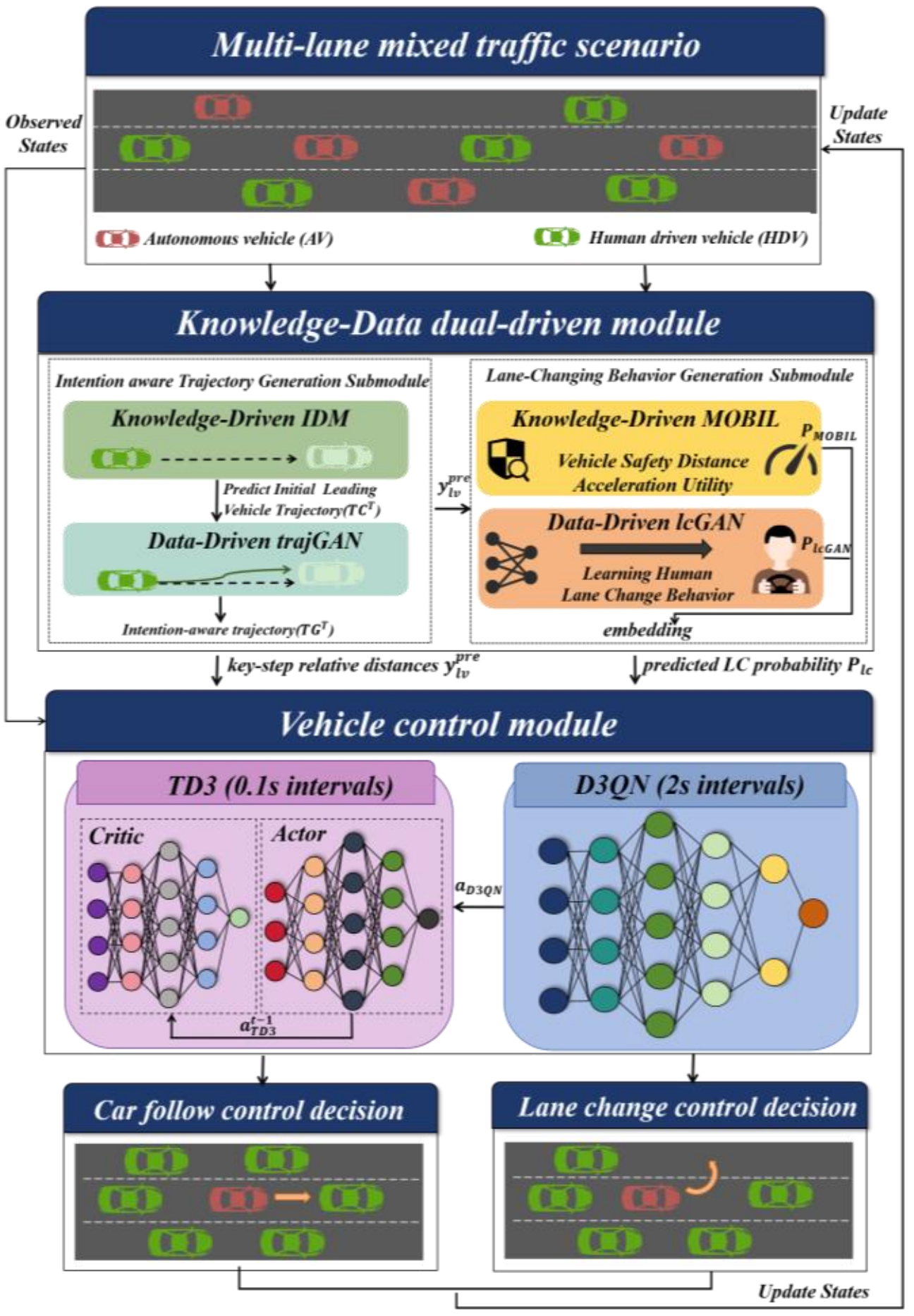

**Fig.2.** Overall framework diagram

### *B. Intention aware Trajectory Generation Submodule*

Conventional car-following and lane-changing decisions typically rely solely on local observations at the current time step, lacking a global understanding of the leading vehicle's behavioral trend, which results in delayed responses and insufficient foresight. To address this issue, this study draws upon the physics-informed deep generative model proposed by Xu et al. (2025) and extends it to the task of multiple leading vehicle trajectories prediction that accounts for driving interaction intentions.

First, we input the speed and position of the leading vehicle into the physical IDM to generate an initial deterministic estimate of the leading vehicle's trajectory serving as a physics-based prior, as shown in **Fig.3**. Subsequently, a data-driven trajGAN is employed to refine this initial trajectory into intention-aware trajectories.

The IDM computes the desired acceleration based on the driving states (speed and position) of the leading vehicle ($lv$):

$$a_{lv}^t = a_{max}\left\{1 - (\frac{v_{lv}^t}{v_0})^\delta - \left[\frac{s^{t*}(v_{slv}^t, v_{lv}^t)}{y_{slv}^t - y_{lv}^t}\right]^2\right\} \quad (11)$$

$$s^{t*}(v_{slv}^t, v_{lv}^t) = s_0 + Tv_{lv}^t + \frac{v_{lv}^t(v_{slv}^t - v_{lv}^t)}{2\sqrt{a_{max}b}} \quad (12)$$

Where $a_{lv}^t$ represents the acceleration of $lv$ at the current time step $t$. The calibrated parameters of IDM are presented in Appendix **Table A.1**.

The physics-based initial trajectory provides a conditioning sequence $TC^T$ for the subsequent trajGAN. On this basis, trajGAN takes $TC^T$ and a Gaussian noise sequence $Z^T \sim N(0, I)$ as inputs, generating a refined intention-aware trajectory $TG^T$:

$$TG^T = G(Z^T, TC^T) \tag{13}$$

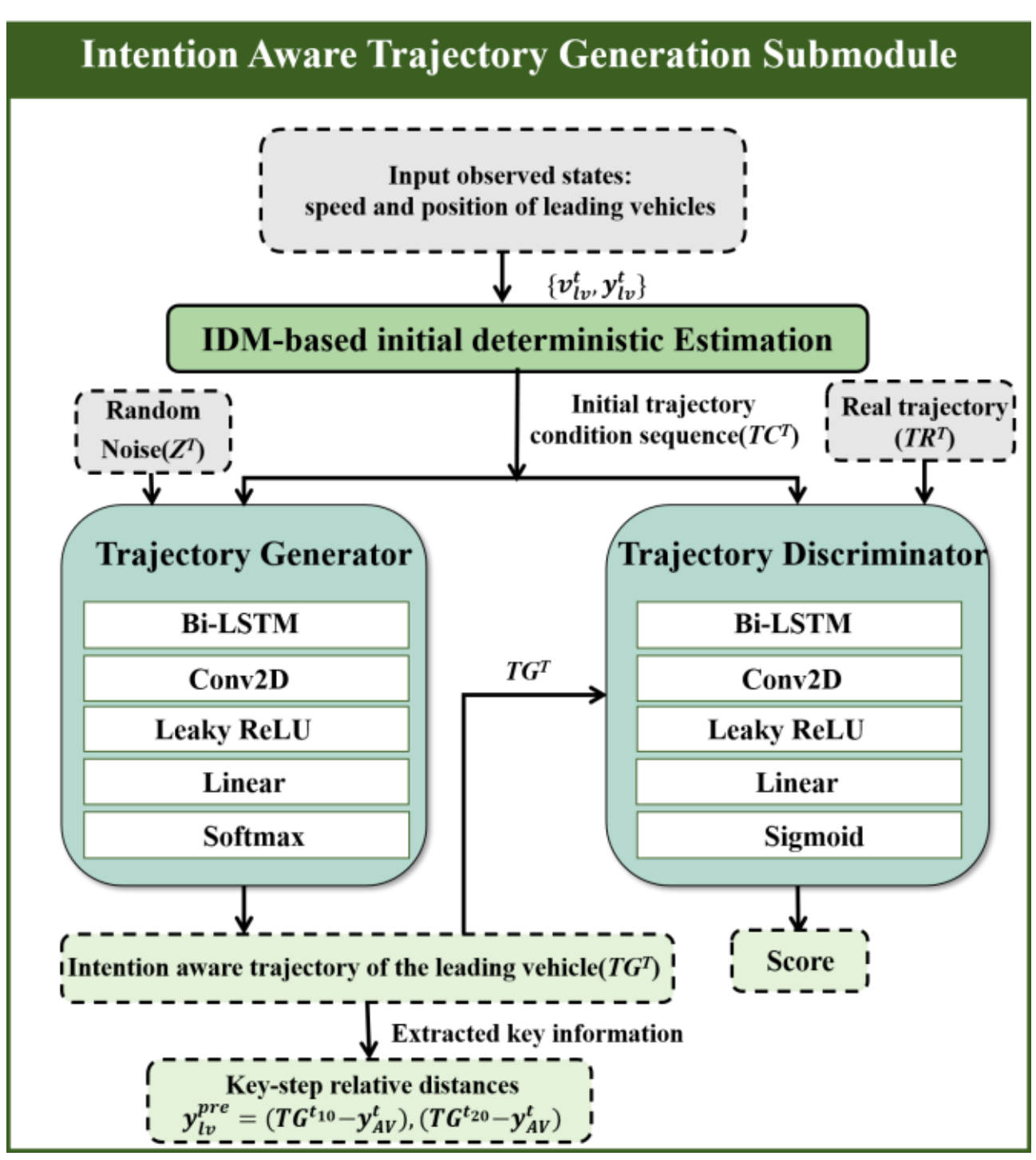

**Fig.3.** The Framework of the Intention aware Trajectory Generation Submodule

The Discriminator $D$ assigns scalar scores to distinguish real trajectory sequence $TR^T$ from generated trajectories $TG^T$. The loss function of the discriminator $L_D^T$ is formulated using binary cross-entropy over mini-batch size $N$:

$$L_D^T = -\frac{1}{N}\sum_{i=1}^{N} \log\left(D(TR_i^T, TC_i^T) + log(1 - D(G(Z_i^T, TC_i^T), TC_i^T)\right) \tag{14}$$

To mitigate the instability of adversarial training and enforce trajectory smoothness, we augment the Generator loss $L_G^T$ with a supervised mean absolute error (MAE) reconstruction loss:

$$L_G^T = -\frac{1}{N}\sum_{i=1}^{N}\{-\log\left(D(G(Z_i^T, TC_i^T), TC_i^T)\right\} + \omega\frac{1}{N}\sum_{i=1}^{N}|TR_i^T - G(Z_i^T, TC_i^T)| \tag{15}$$

where $\omega$ is the reconstruction loss weight. Through the jointing optimization of adversarial and reconstruction losses, trajGAN learns real-world behavioral distributions and generates physically plausible trajectories encoding interactive driving intentions.

This study employs trajGAN to predict 2-second trajectories (20-time steps at 0.1s resolution) for the leading vehicles in the AV's current lane and adjacent lanes. To prevent high-dimensional state representation explosion in downstream RL, key temporal step are extracted from the 20-step prediction. Specifically, key relative distances($y_{lv}^{pre}$)at the 10th step $(TG^{t_{10}} - y_{AV}^t)$ and 20th step $(TG^{t_{20}} - y_{AV}^t)$ relative to the AV's current position($y_{AV}^t$) are computed. These compact relative distances are passed to vehicle control module, providing structured foresight without compromising learning stability. In this manner, the generated trajectories capture both the motion patterns of the leading vehicles and their interaction intentions with the AV, thereby providing forward-looking environmental awareness for safe car-following and lane-changing decisions.

### C. Lane-Changing Behavior Generation Submodule

The framework of lane-changing behavior generation submodule, illustrated in **Fig.4**, comprises of the physical MOBIL model and the data-driven model lcGAN. The input for both MOBIL and lcGAN consists of the current traffic state from environmental perception including the speeds, accelerations, and relative positions of the AV and its surrounding vehicles, as well as the key-step relative distances calculated by the intention-aware trajectory generation submodule. lcGAN learns human lane-changing behavior patterns from real-world datasets and generates a data-driven lane-changing probability distribution $P_{lcGAN}$, based on the empirical maneuver distribution. Furthermore, MOBIL calculates the utility value of each lane-changing action through safety distance checking and efficiency utility computation, outputting a knowledge-driven lane-changing probability distribution $P_{MOBIL}$. Finally, the data-driven and knowledge-driven lane-changing probabilities are compressed and integrated via an embedding layer to form a comprehensive prior distribution $P_{lc}$, which is fed into the downstream RL model to guide the lane-changing decision process.

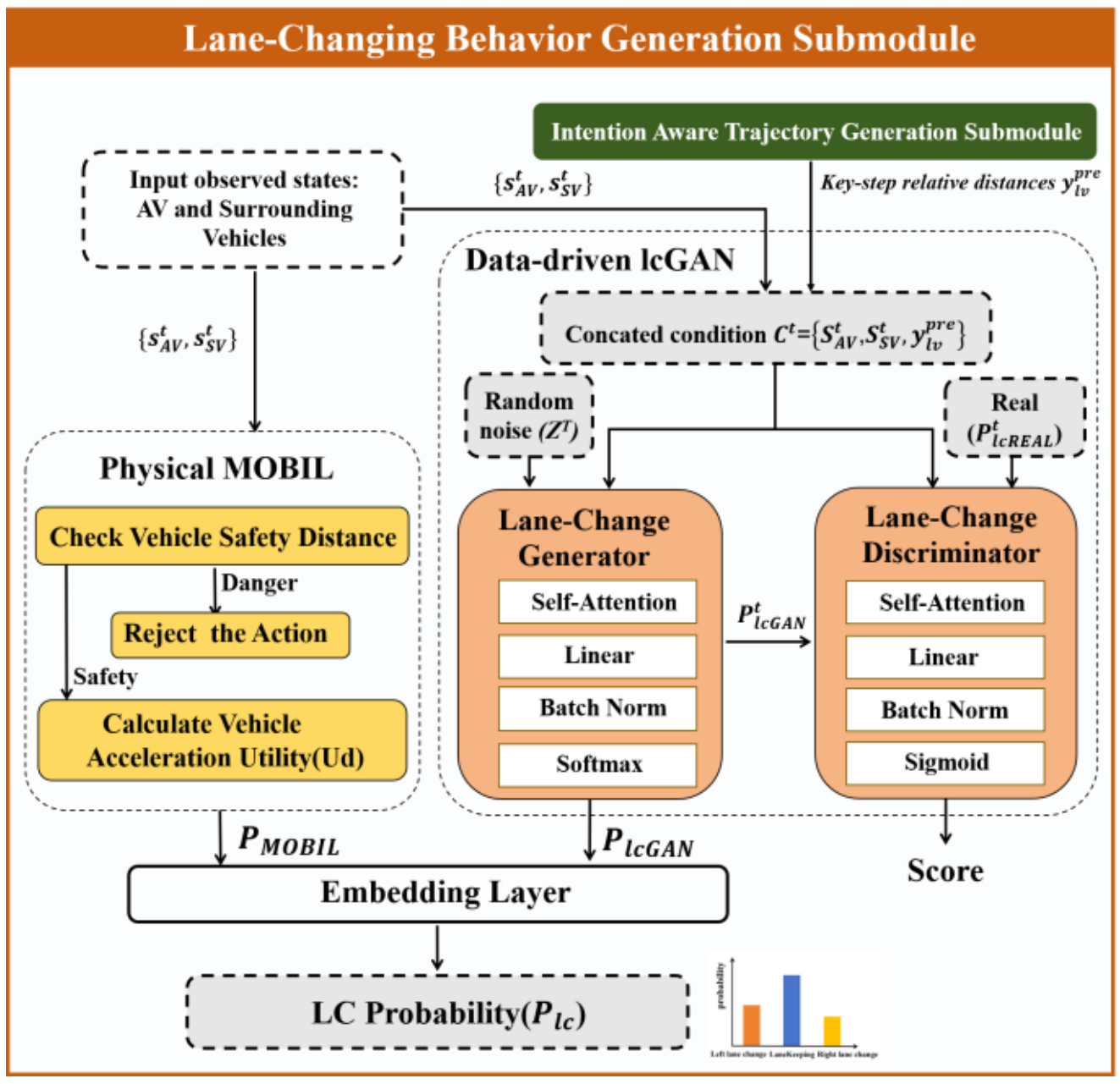

**Fig. 4.** The framework of lane-changing behavior generation submodule

#### 1) Data-Driven lcGAN

In the context of vehicle lane-changing decision making, traditional RL-based policies often suffer from non-stationarity during safety-critical interactions as surrounding vehicles, causing distribution shift and leaving long-tail safety events under-explored. To overcome these issues, this paper designs lcGAN, a generative model that extracts lane-changing decision patterns from empirical driving data to generate a data-driven probability distribution conditioned on the intention-aware trajectories and prevailing traffic state. This provides an informative auxiliary prior to the RL agent for more efficient and safer decision-making. Employing a GAN structure to generate lane-change probabilities explicitly captures individual driver behavior heterogeneity.

The input to the Generator $G$ contains two parts: 1) a noise vector $Z^t \sim N(0,1)$ sampled from a standard normal distribution; 2) a condition vector $C^t$, consisting of the observed current traffic states of the AV ($S_{AV}^t$) and its surrounding vehicles ($S_{SV}^t$), as well as the key-step relative distances $y_{lv}^{pre} = [(TG^{t_{10}} - y_{AV}^t), (TG^{t_{20}} - y_{AV}^t)]$ calculated from the intention-aware trajectory generation submodule. These condition features are concatenated into a condition vector $C^t$=$\{S_{AV}^t, S_{SV}^t, y_{lv}^{pre}\}$. lcGAN generates the data-driven lane-changing probability distribution $P_{lcGAN}^t$ at the time step *t*, which includes the probabilities of keeping the current lane, changing to the left lane, or changing to the right lane: $P_{lcGAN}^t$=$\{P_{left}, P_{keep}, P_{right}\}$. The probability of each direction ranges from 0 to 1, with their summation constraining to 1:

$$P_{lcGAN}^t = G(Z^t, C^t) \tag{16}$$

The Discriminator $D$ evaluates whether an input lane-changing probability vector represents authentic human driving behavior or generated samples. Its input $P_{lc}$ includes either the generated probability $P_{lcGAN}^t$ or the real lane-changing action $P_{lcREAL}^t$ (one-hot encoded as [1,0,0] for left change; [0,1,0] for lane keeping; [0,0,1] for right change, conditioned on $C^t$:

$$s = D(P_{lc}, C^t) \tag{17}$$

where $s \in [0,1]$ represents a scalar score output by the Discriminator $D$, indicating the probability that the input lane-changing distribution $P_{lc}$ corresponds to an real human driving behavior or generated sample, given the condition vector $C^t$.

An adversarial training approach integrating both adversarial and supervised losses is adopted. The loss functions of the discriminator $L_d^t$ and the generator $L_g^t$ over mini-batch size $N$ are expressed as:

$$L_d^t = -\frac{1}{N}\sum_{i=1}^{N}\{\log\left(D(P_{lcREAL,i}^t, C_i^t) + log(1 - D(G(Z_i^t, C_i^t), C_i^t))\right\} \tag{18}$$

$$L_g^t = -\frac{1}{N}\sum_{i=1}^{N}\log\left(D(G(Z_i^t, C_i^t), C_i^t)\right) + \omega\frac{1}{N}\sum_{i=1}^{N}\left(-\alpha_k(1-p_{i,k})^{\gamma} log(p_{i,k})\right) \tag{19}$$

$L_g^t$ includes the adversarial loss and supervised loss of the generator, and $\omega$ is the supervised loss weight. Given the scarcity of lane-changing maneuvers relative to lane-keeping in naturalistic traffic data, a customized supervised loss (Focal loss) is employed to mitigate class imbalance by down-weighting easy lane-keeping examples and focusing on rare lane-changing actions (Lin et al.,2020). Here $\alpha_k$ denotes the class balance weights for action $k \in \{\text{left}, \text{keep}, \text{right}\}$, $p_{i,k}$ is the predicted probability for the ground-truth action class, and the focusing parameter $\gamma$ is set to 2 according to the empirical calibration.

### 2) *Knowledge driven MOBIL*

Pure end-to-end RL methods face key limitations in practical applications: (a) the neural network decision-making process lacks explicit physics-informed guidance, potentially leading to dangerous maneuvers that violate traffic dynamics (e.g., frequent lane changes or hard acceleration/deceleration); (b) unconstrained exploration in the early training stage is sample-inefficient, resulting in delayed convergence. To address these issues, this paper introduces a physical knowledge-driven model, MOBIL. The built-in safety checking mechanism of the MOBIL enforces hard safety distance constraints to prevent the RL agent from attempting dangerous lane-changing maneuvers during exploration, while its physics-based incentive function provides efficiency guidance, avoiding ineffective maneuvers and offering a physically interpretable prior.

MOBIL evaluates lane-changing decisions based on safety and efficiency criteria. Safety distance checking is a prerequisite. For the preceding ($TL$) and following vehicles ($FT$) in the candidate target lane, safety constraints must be satisfied:

$$\begin{cases} y_{TL} - y_{AV} > y_{safe} \\ y_{AV} - y_{FT} > y_{safe} \end{cases} \tag{20}$$

where $y_{AV}$,$y_{TL}$, and $y_{FT}$ are the longitudinal positions of the AV, the leading vehicle in the target lane, and the following vehicle in the target lane, respectively; $y_{safe}$ denotes the minimum safe gap.

If safety constraints in Equ.(20) are not satisfied, the corresponding lane-changing probability is set to 0. If safety constraints are satisfied, the utility value of the action $U_d$ for direction $d \in \{left, right, keep\}$ is calculated via Equ.(21):

$$\begin{cases} U_d = (a_{AV}^{new} - a_{AV}^{old}) + p[(a_{CF}^{new} - a_{CF}^{old}) \\ \quad +(a_{TF}^{new} - a_{TF}^{old})] > \Delta a_{th} \\ \quad a_{TF}^{new} > -b_{safe} \end{cases} \tag{21}$$

where $a_{AV}^{new}$ and $a_{AV}^{old}$ are the accelerations of the AV before and after the potential lane change; $a_{CF}^{new}$and $a_{CF}^{old}$ are the accelerations of the following vehicle in the current lane; $a_{TF}^{new}$ and $a_{TF}^{old}$ are the accelerations of the following vehicle in the target lane. The parameters of MOBIL are shown in Appendix **Table A.1.**

To adapt to the D3QN discrete action space, the utility values calculated by MOBIL are converted into a normalized probability distribution $P_{MOBIL} = [p_{left}, p_{right}, p_{keep}]$:

$$P_{MOBIL} = \begin{cases} p_{left} = \dfrac{\max(0, U_{left})}{\sum_d \max(0, U_d)} \\ p_{right} = \dfrac{\max(0, U_{right})}{\sum_d \max(0, U_d)} \\ p_{keep} = 1 - p_{left} - p_{right} \end{cases}, d \in \{left, right, keep\} \tag{22}$$

where $U_{keep} > 0$ denotes a preset baseline utility for maintaining the current lane, ensuring a non-zero probability for lane keeping when lane-changing incentives are low. $U_{keep}$ is calculated by inferring the acceleration of AV and its following vehicle through the IDM in the current lane. If $U_d \le \Delta a_{th}$, the probability of that action is set to 0. Finally, the normalized probability distribution $P_{MOBIL}$ across all three discrete actions is calculated using Equ.(22).

### 3) *Integration of Knowledge-Driven and Data-Driven Modules*

The data-driven lcGAN extracts empirical lane-changing patterns from real-world driving data, yielding $P_{lcGAN}$. Conversely, MOBIL enforces physics-based safety distances and utility boundaries, supplying inductive biases $P_{MOBIL}$. To capitalize on their complementary strengths without causing state dimensionality explosion in the RL agent, an embedding-based fusion layer maps $P_{lcGAN}$ and $P_{MOBIL}$ into a compact, low-dimensional prior representation $P_{lc}$:

$$P_{lc} = embedding(P_{lcGAN}, P_{MOBIL}) \tag{23}$$

This fusion preserves critical safety constraints and empirical maneuver trends while projecting them into a compact vector $P_{lc}$.

*G. Vehicle Control Module*

The framework of vehicle control module, illustrated in **Fig. 5**, adopts a hierarchical RL architecture operating across different time scales. The inputs to the module is constructed by raw traffic observations from environmental perception, including AV state (longitudinal and lateral speeds, longitudinal position) and surrounding vehicles SV state (relative speeds and positions with respect to the AV), with the predictive outputs generated by the preceding submodules. These predictive outputs include key-step relative distance $y_{lv}^{pre}$ from the intention-aware trajectory generation submodule and the fused prior distribution $P_{lc}$ from the lane-changing behavior generation submodule. The vehicle control module executes differentiated RL policies tailored to heterogeneous action spaces: lane-changing decisions are handled by a low-frequency D3QN agent every 2 seconds, while car-following control is handled by a high-frequency TD3 agent every 0.1 seconds. This asynchronous design aligns with naturalistic human driving logic.

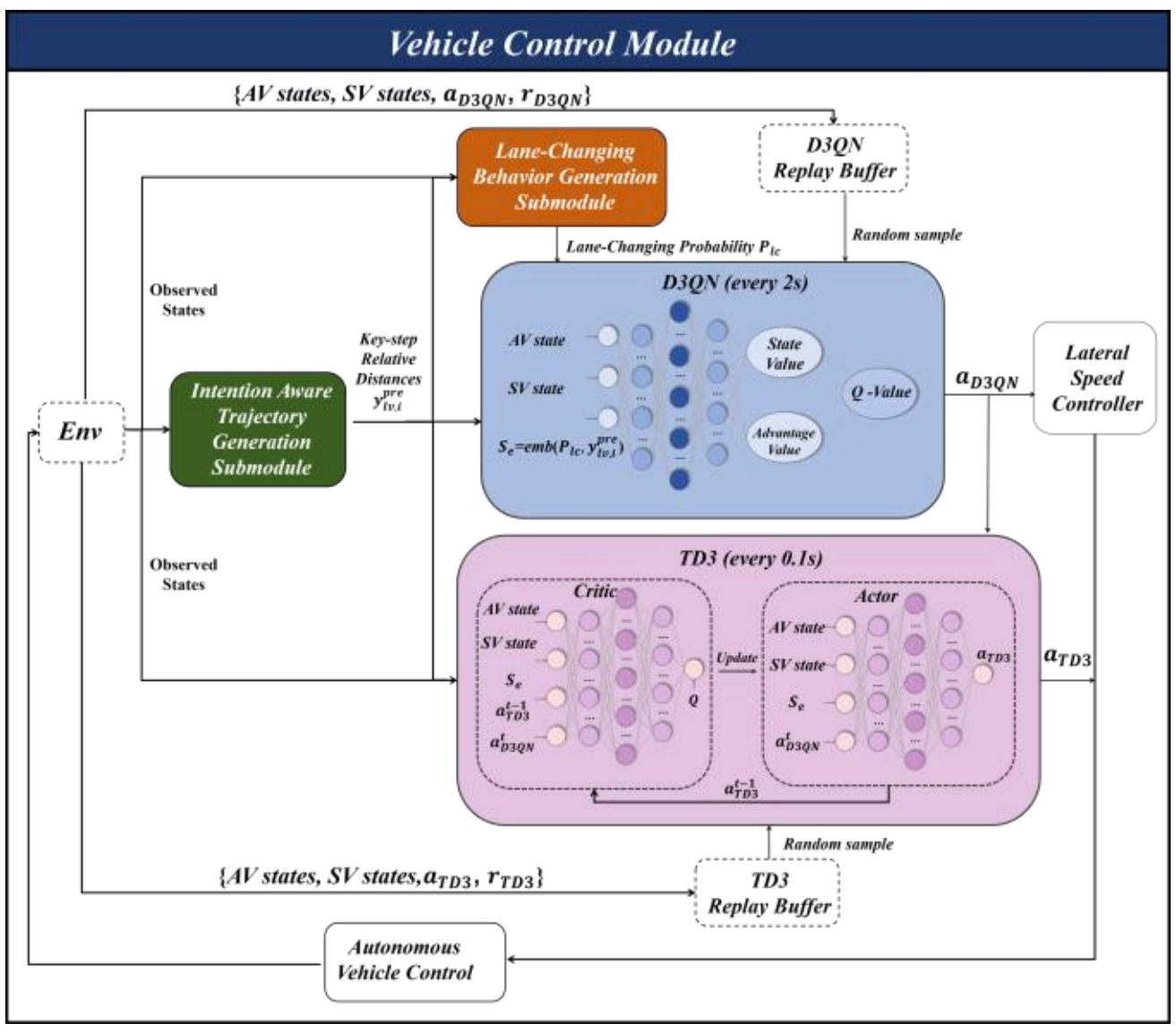

**Fig.5.** The framework of the vehicle control module

***1) Lane-Changing Decision Based on the D3QN Model***

This paper employs the D3QN model to address the discrete lane-changing decision problem of AV in mixed traffic flow. By decoupling action-value function into the state-value function and advantage function, D3QN effectively evaluates the long-term benefits of different actions, thereby enhancing decision stability and learning efficiency in complex stochastic environments. This study enriches the traditional observed vehicle state representation to enhance its anticipatory environmental perception and decision-making stability. The improvements are reflected in the following two key aspects: first, the intention-aware trajectory information of leading vehicles is incorporated (Section III.B); second, knowledge-data dual driven prior lane-change distribution incorporating both empirical data distribution and physical safety constraints is integrated (Section III.C). To illustrate the input state and action space, a mixed traffic scenario is constructed as in **Fig. 6**, where the ego AV (red one) is controlled by the proposed framework for both lane-changing and car-following maneuvers, while the remaining vehicles (green ones) are surrounding HDVs labeled as a, b, c, d, e and f.

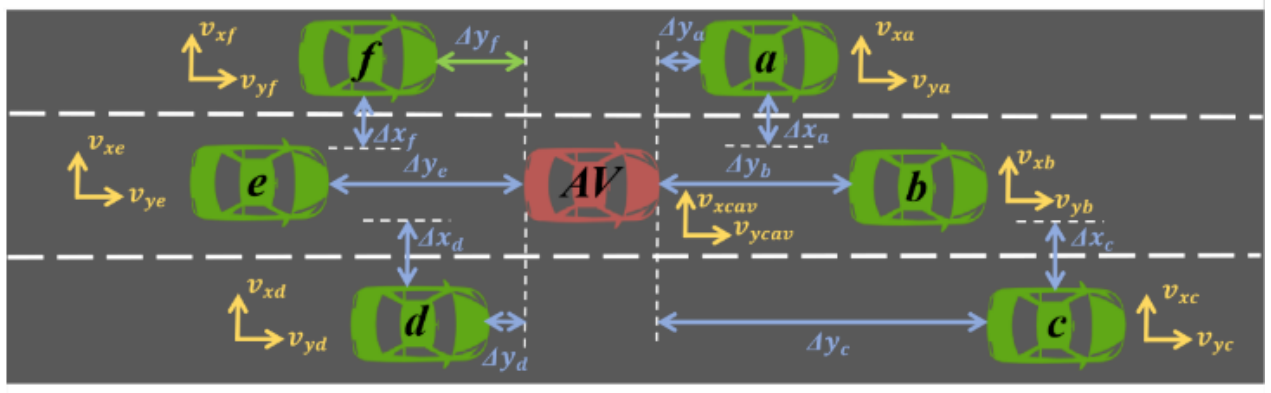

**Fig.6.** Mixed traffic flow scenario

**(1) State Space:** the state space $s_{D3QN}$ includes the observed state of AV ($s_{AV}$), the relative states of surrounding vehicles ($\Delta s_i$), and a unified embedded state $s_e$ combining intention-aware trajectories and knowledge-data dual-driven lane change priors via a shared embedding layer:

$$s_{D3QN} = [s_{AV}, \Delta s_i, s_e]$$
$$s_{AV} = [y_{AV}, v_{AV,y}, v_{AV,x}]$$
$$\Delta s_i = [\Delta x_i, \Delta y_i, \Delta v_{x,i}, \Delta v_{y,i}], i \epsilon (a, b, c, d, e, f)$$
$$s_e = embedding(P_{lc}, y_{lv,i}^{pre})$$
$$y_{lv,i}^{pre} = [(TG_i^{t_{10}} - y_{AV}^t), (TG_i^{t_{20}} - y_{AV}^t)], i \in (a, b, c) \quad (24)$$

where $y_{AV}, v_{AV,y}$ and $v_{AV,x}$ denote the observed longitudinal position, longitudinal velocity, and lateral velocity of the AV, respectively; $\Delta s_i$ denotes the observed relative spatial and kinematic state of surrounding vehicle $i$ relative to AV, including relative lateral and longitudinal distance ($\Delta x_i$ and $\Delta y_i$), relative lateral and longitudinal velocity ($\Delta v_{x,i}$ and $\Delta v_{y,i}$). $y_{lv,i}^{pre}$ denotes the key-step relative distance calculated from intention-aware trajectories of leading vehicles a, b and c.

**(2) Action Space:** the action space for D3QN $a_{D3QN}$ is discrete, comprising three decision choices: left lane change, lane keeping, and right lane change.

$$a_{D3QN} = \{Left, Keep, Right\} \quad (25)$$

**(3) Reward Function:** the reward of D3QN $r_{D3QN}$ is calculated at each 2-second decision step to jointly optimize driving safety, efficiency, and comfort. An efficiency reward $r_{eff}$ encourages the AV to maintain a high longitudinal speed within the speed limit, explicitly linking speed to reward and thus promoting timely lane changes when the desired speed is constrained in the current lane. A lane-changing penalty $r_{lc}$ imposes a persistent cost on every lane-change action to suppress frequent or unnecessary maneuvers that degrade comfort. A collision penalty $r_{collision}$ is set to a large negative value, immediately terminating the current training episode to enforce safety imperative. The overall reward structure is formulated as:

$$r_{D3QN} = r_{eff} + r_{lc} + r_{collision} \quad (26)$$

$$r_{eff} = \begin{cases} \frac{v_{AV,y} - v_{min}}{v_{max} - v_{min}} & if\ v_{min} \le v_{AV,y} \le v_{max} \\ -1 & otherwise \end{cases} \quad (27)$$

$$r_{lc} = \begin{cases} -1 & if\ lane\ change \\ 0 & otherwise \end{cases} \quad (28)$$

$$r_{collision} = \begin{cases} -100 & if\ collision \\ 0 & otherwise \end{cases} \quad (29)$$

where $v_{AV,y}$ is the longitudinal speed of the AV, while $v_{max}$ and $v_{min}$ are the upper and lower speed limits of the roadway, respectively.

***2) Car-Following Decision Based on the TD3 Model***

The TD3 algorithm is adopted to handle the continuous car-following control problem due to its suitability for continuous action spaces and its ability to mitigate Q-value overestimation bias, leading to stable and reliable policy learning. The state representation of TD3 is augmented by incorporating intention-aware trajectories, enabling the agent to anticipate the future motion trend of leading vehicles when executing continuous acceleration decisions, thereby achieving smoother, safer, and more forward-looking car-following decisions.

**(1) State Space:** the state space of the TD3 $s_{TD3}$ contains observed kinematic states of AV ($s_{AV}$), relative states of surrounding vehicle ($\Delta s_i$), the unified embedded state ($s_e$), the discrete lane-changing action output by D3QN ($a_{D3QN}^{t}$) and the continuous acceleration action executed at the previous time step $a_{TD3}^{t-1}$:

$$s_{TD3} = \left[s_{AV}, \Delta s_i, s_e, a_{D3QN}^{t}, a_{TD3}^{t-1}\right] \tag{30}$$

where $s_{AV}, \Delta s_i$ *and* $s_e$ share identical formulation with the D3QN state representation. Because TD3 operates at a higher control frequency (every 0.1s) than D3QN (every 2s), the lane-changing action $a_{D3QN}^{t}$ remains persistent across 20 consecutive TD3 control steps within one D3QN decision interval.

**(2) Action Space:** the action space $a_{TD3}$ is a continuous one-dimensional acceleration command. Considering physical acceleration/deceleration constraints on roadways, the allowable acceleration action range is bounded as (Ni et al.,2024):

$$a_{TD3} \in \left[-5\, m/s^2, 5\, m/s^2\right] \tag{31}$$

**(3) Reward Function:** the reward $r_{TD3}$ is calculated at each 0.1-second decision step to jointly optimize safety, efficiency, and comfort during car-following maneuvers. The overall reward structure is formulated as:

$$r_{TD3} = r_{eff} + r_{acc} + r_{collision} \tag{32}$$

The efficiency reward $r_{eff}$ and $r_{collision}$ follow the definitions in (27) and (29). To ensure acceleration comfort and avoid excessive inertial jerk, an acceleration penalty $r_{acc}$ is introduced. Fu et al. (2020) indicates that acceleration exceeding 4 m/s² significantly degrade driving comfort. Thus, $r_{acc}$ penalizes extreme acceleration/deceleration exceeding this threshold:

$$r_{acc} = \begin{cases} -(|a_y| - 4)^2 & if \ |a_y| > 4\, m/s^2 \\ 0 & otherwise \end{cases} \tag{33}$$

### 3) *Hierarchical Co-Training and Execution Framework*

Consequently, the proposed KDDRL framework accomplishes integrated training of lane-changing and car-following through hierarchical decision-making: the upper-level D3QN agent executes low-frequency (every 2s) discrete lane-changing decisions guided by a prior distribution $P_{lc}$ that fuses data-driven lane-changing probabilities from lcGAN and the knowledge-driven safety utility from MOBIL; the lower-level TD3 agent handles high-frequency (every 0.1s) continuous car-following control, incorporating trajGAN-predicted intention aware trajectories into its state space for proactive vehicle control.

The two agents operate cooperatively in a shared simulation environment across distinct time scale, learning asynchronously through their respective reward functions and independent replay buffers. The D3QN is updated every 20 simulation steps (2s), whereas the TD3 is updated at every simulation step (0.1s). The detailed training procedure is presented in Appendix **Algorithm 1**. An episode terminates if a collision occurs, the vehicle goes out of bounds, or the current episode reaches the maximum of 1000 steps.

## IV. EXPERIMENTS

### A. *Data Preparation*

To enhance the realism of the simulation environment, this study builds a simulation environment for training and testing the proposed framework by integrating real-world trajectory data from the HighD dataset in to SUMO simulation platform (Lopez et al.,2018). The HighD dataset collects vehicle trajectory data on German highways using drone-mounted aerial equipment, as shown in **Fig. 7**. Recorded at a high frame rate of 25 Hz, the dataset comprises a total of 60 recordings, with an average duration of 17 minutes and an average road coverage length of 420 meters. Advanced computer vision techniques extract key vehicle state parameters, including vehicles' lateral and longitudinal positions, speeds, accelerations, and driving directions (Krajewski et al.,2018).

In this study, naturalistic vehicle trajectories extracted from the HighD dataset are employed to calibrate the driving maneuver parameters of the surrounding human-driven vehicles in the simulation, ensuring that background traffic exhibits realistic driving maneuver. Specifically, the parameter sets of the SUMO's built-in car-following model (IDM) and lane-changing model (MOBIL) are calibrated using the HighD trajectory data (calibrated values provided in Appendix **TABLE A.1**). During simulation, the surrounding HDVs are controlled by these calibrated IDM and MOBIL models, while the ego AV is governed by the proposed KDDRL framework. The experiments were conducted on a computer equipped with two Hygon C86-3G processors (3.10 GHz, OPN: 3390) and an NVIDIA GeForce RTX 4090 graphics card.

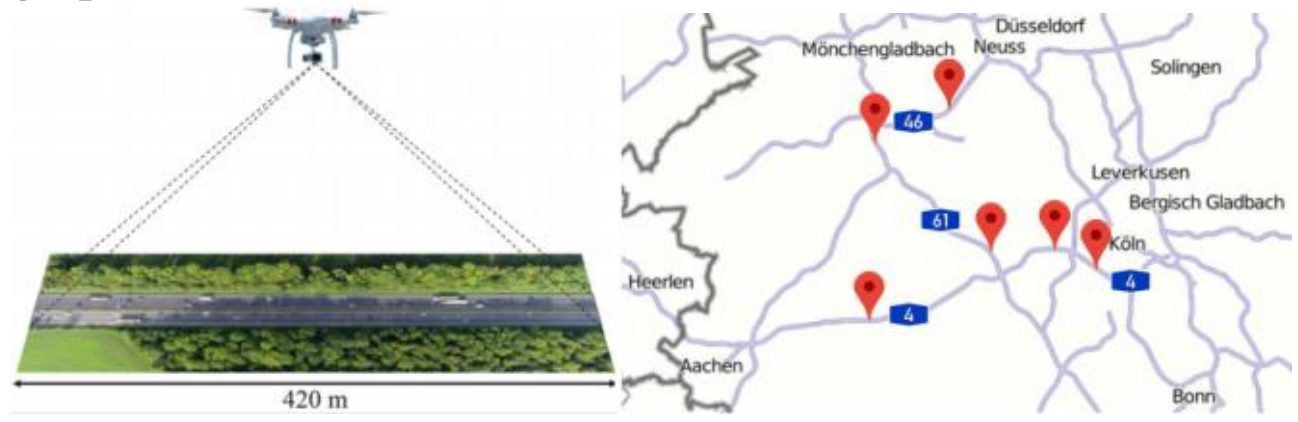

**Fig.7.** Configuration and locations of the HighD dataset

### B. *Training Setup*

Taking a west-to-east one-way three-lane scenario of a German highway around Cologne as the study scenario, a total of 2,325 vehicles' real trajectory data are extracted over one-hour period, wherein 15.8% of the vehicles perform a lane change maneuvers. The complete traversal of a vehicle from the entry to the exit of the study area is considered as one episode. The decision frequency of D3QN is set to 2s, and that of TD3 is set to 0.1s. This design aligns with the low-frequency nature of human lane-changing decisions while ensuring operational continuity and responsiveness for continuous car-following control. The simulation time step in SUMO is uniformly set to 0.1s.

During training, the trajGAN and lcGAN models within the Knowledge-Data dual-driven module are pre-trained offline using the observed trajectories to capture interaction intentions and human maneuver patterns. Subsequently, the parameters of these pre-trained generative models are then integrated into the vehicle control module to further train the D3QN and TD3 models.

### C. *KDDRL Model Training Results*

This paper compares the proposed KDDRL model with four recent state-of-the-art RL baselines and a classic rule-based physical model, as shown in **TABLE** I. All the baselines are trained for 1000 episodes under identical environmental conditions. To ensure experimental consistency and unbiased evaluation, one vehicle is randomly designated as the controlled ego AV in each episode.

TABLE I
KDDRL VERSUS THE BASELINES

| Algorithm | Lane change model | Car-follow model | Fram e-work | Control frequen-cy | Physics features | Data-driven feature |
|---|---|---|---|---|---|---|
| Physical model | MOBIL | IDM | Hier | (0.1s) | MOBIL +IDM | / |
| Ni et al. | DDQN | TD3 | Hier | (0.1s / 1.0s) | / | / |
| Peng et al. | D3QN | DDPG | Hier | (0.1s) | MOBIL | / |
| Lin et al. | D3QN | D3QN | Uni | (0.1s) | / | / |
| Cui et al. | PASAC | PASAC | Uni | (0.1s) | / | / |
| KDDRL | D3QN | TD3 | Hier | (0.1s / 2.0s) | MOBIL +IDM | trajGAN +lcGAN |

Note: Hier and Uni meas hierarchical and unified framework seperately.

The convergence of cumulative rewards for the lane-changing and car-following control during training is shown in **Figs. 8(a)** and **8(b)**. All compared baselines retain their original structures and parameters as specified in their respective publications. Due to differences in environment settings and the inherent randomness in policy exploration, the reward curves exhibit substantial fluctuations during the initial training phase. As the number of training episodes increases, the policies of all methods gradually stabilize. Compared with the other RL baselines, the proposed KDDRL framework achieves faster convergence to the optimal policy and attains the highest cumulative reward.

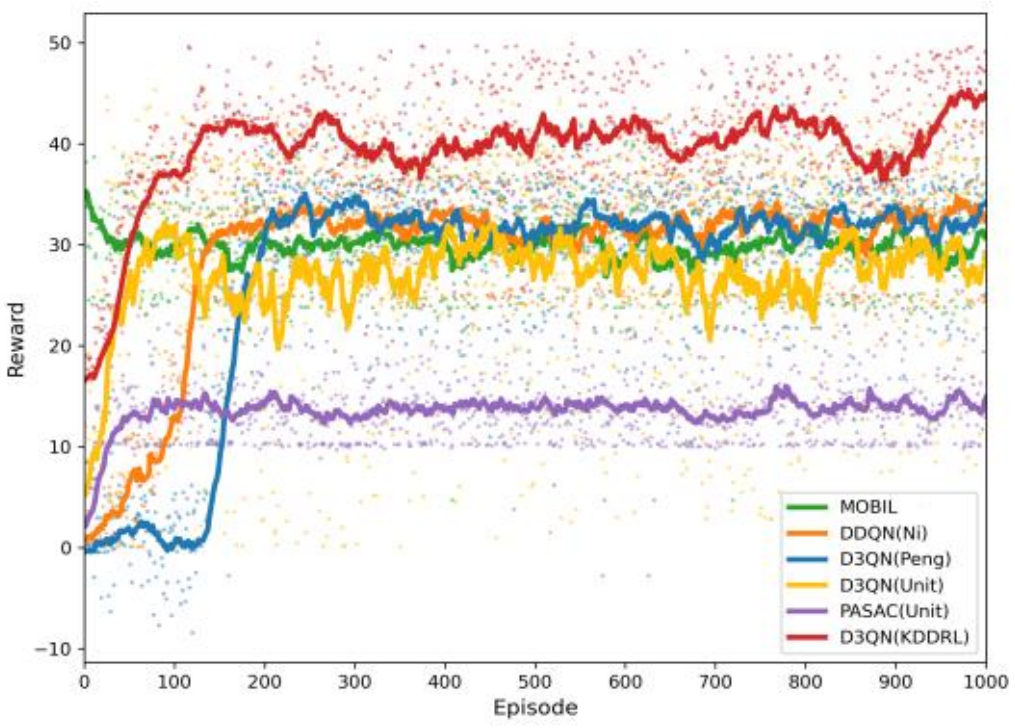

(a)lane-changing control

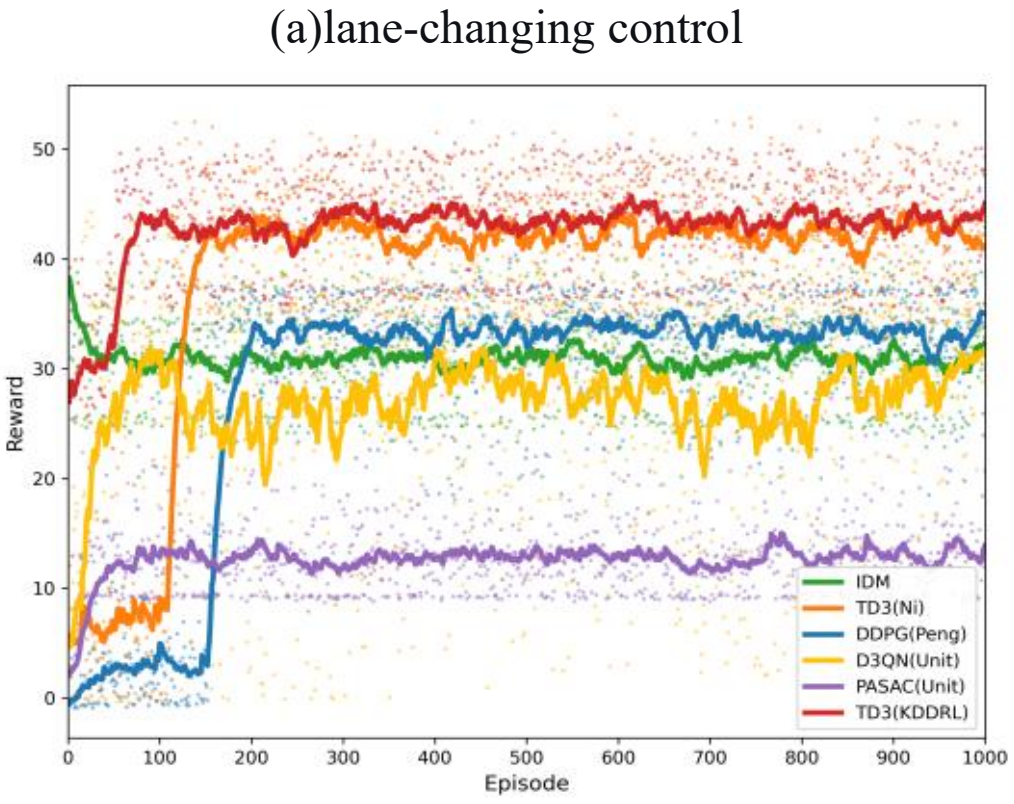

(b)car-following control

**Fig.8.** Reward distribution curves during training

**Fig. 9** presents box plots of the average reward over the course of training. The proposed KDDRL (shown in red), which integrates hierarchical D3QN and TD3, demonstrates superior mean and median reward values, verifying the decision-making stability and learning efficiency of the framework in complex traffic scenarios. Furthermore, the reward trends reveal that the unified control frameworks (shown in yellow and purple) yield lower rewards than the hierarchical control frameworks in both lane changing and car following tasks. This can be attributed to the fact that the handling hybrid action spaces within a unified frameworks destabilize the optimization process, while a naive decoupling without structured coordination fails to capture longitudinal-lateral interdependencies.

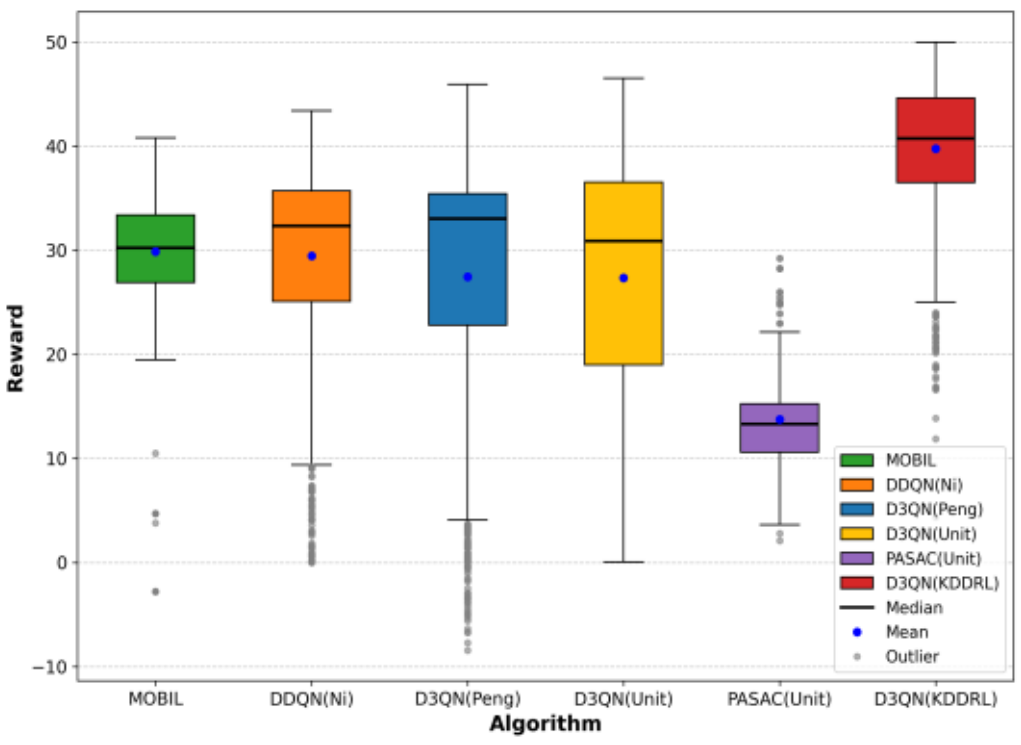

(a)lane-changing control

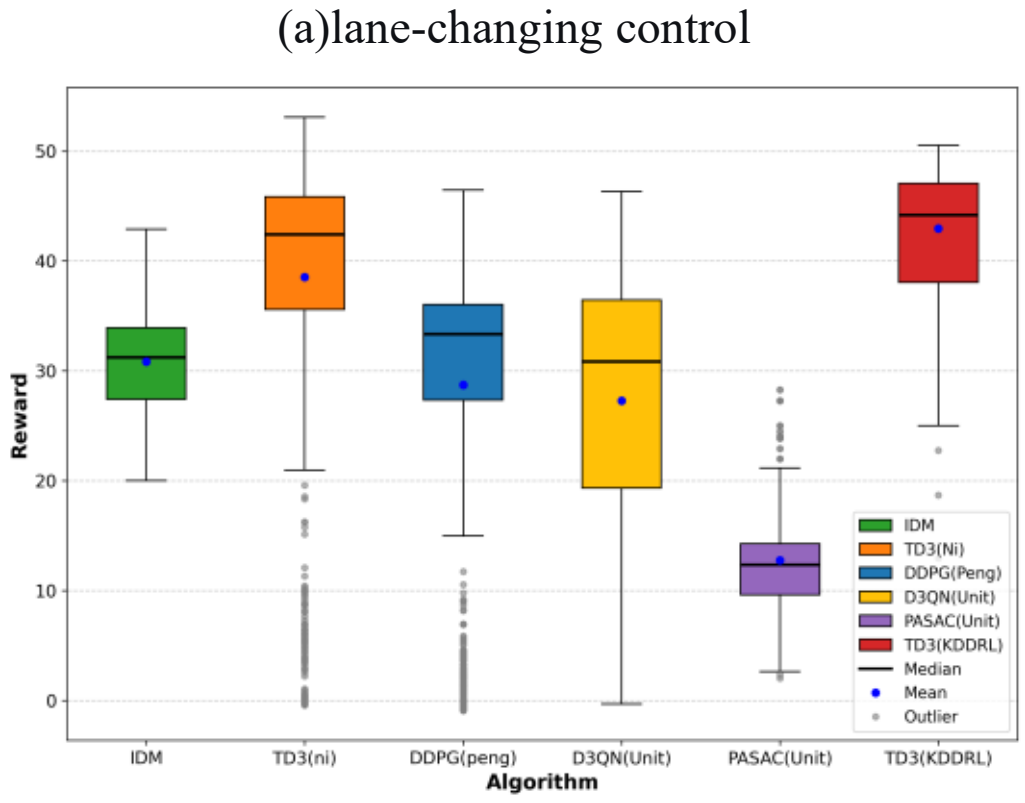

(b)car-following control

**Fig.9** The box plot of the average reward during training

To further verify the decision-making performance of the proposed method, the converged polices were evaluated under identical simulation scenarios. **Fig. 10** presents the speed and acceleration distributions of the ego AV under different control methods, where shaded regions indicate variance across multiple test runs. **TABLE II** summarizes the quantitative performance metrics across the five methods. The evaluation results indicate that the average speed of the KDDRL framework exceeds that of all baselines, indicating that the controlled ego AV achieves higher traffic efficiency. The acceleration distribution exhibits a low dispersion (shown in red shaded area in **Fig. 10b**), reflecting that the model's acceleration control is more proactive and smoother, promptly adapting to changes in the traffic environment while maintaining driving comfort.

Regarding safety and efficiency trade-offs, KDDRL achieves optimal performance across the evaluated key metrics, delivering the highest average speed alongside the lowest collision rate. Moreover, the MOBIL-integrated hierarchical framework by Peng et al. shows more frequent lane changes and a higher conflict rate than the pure RL counterpart by Ni et al.. In contrasts, KDDRL balances safety and efficiency by leveraging knowledge-data dual-driven priors and intention-aware trajectory foresights. Notably, compared with hierarchical framework, the unified framework, D3QN(Unit) and PASAC(Unit), yield a much lower average speed while inducing an excessive lane changes. This demonstrates that monolithic architectures tend to generate redundant lateral maneuvers and struggles to coordinate multi-timescale longitudinal and lateral decisions efficiently.

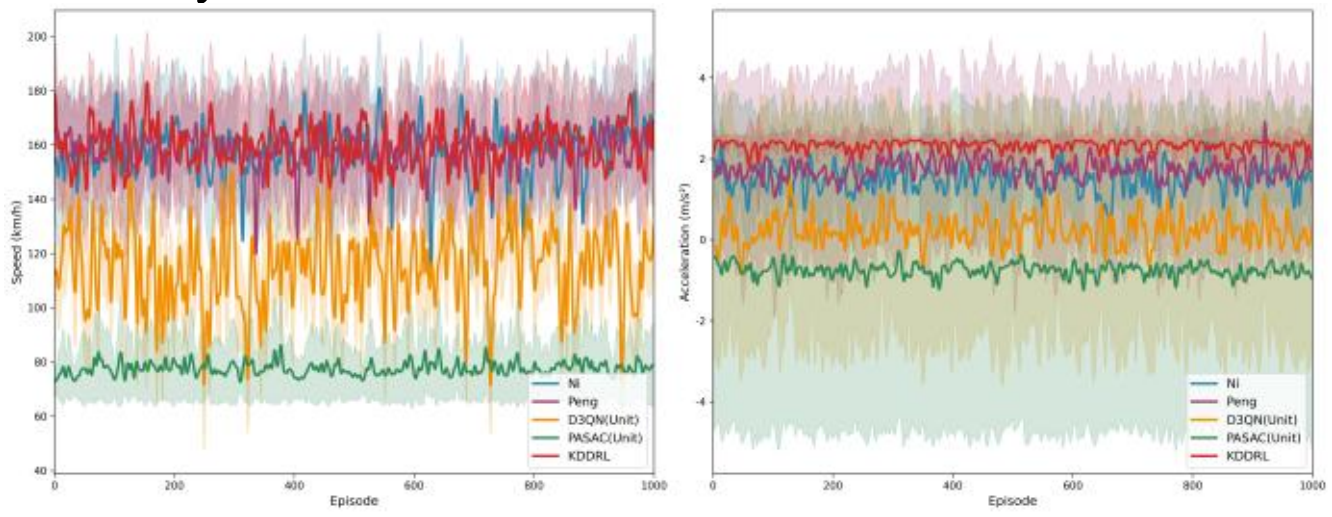

(a) Speed distribution curve (b) Acceleration distribution curve

**Fig.10.** Speed and acceleration distributions of different algorithms during testing

TABLE II
PERFORMANCE COMPARISON OF DIFFERENT ALGORITHMS DURING TESTING

| Algorithm | Average speed (km/h) | Average acceleration (m/s²) | Number of lane changes | Collisi-on rate |
|---|---|---|---|---|
| DDQN+TD3(Ni) | 157.7 ± 23.8 | 1.49 ± 1.0 | 492 ± 17 | 1.4% |
| D3QN+DDPG (Peng) | 158.6 ± 17.8 | 1.75 ± 2.3 | 1778 ± 39 | 2.7% |
| D3QN(Unit) | 115.0 ± 16.0 | 0.22 ± 2.8 | 2175 ± 45 | 2.1% |
| PASAC(Unit) | 77.6 ± 12.0 | -0.75 ± 4.0 | 1709 ± 62 | 0.7% |
| D3QN+TD3 (KDDRL) | **159.4 ± 22.5** | **2.33 ± 0.3** | **1243 ± 29** | **0.1%** |

### *D. Comparison Results in Different Scenarios*

To evaluate the generalization ability and operational robustness of the model in complex traffic environments, this paper designs two additional test scenarios.

**Scenario 1:** Simulates the model's performance under high-density traffic conditions by increasing traffic flow by approximately 50% relative to the baseline (the volume was increased from the observed condition of 886 veh/h/lane to 1329 veh/h/lane). This scenario is shown in **Fig.11(b)**, where the red vehicle represents the ego AV, the green vehicles represent the initial surrounding vehicles, and the blue vehicles represent the added vehicles. Rule-based models (i.e., IDM and MOBIL) are used to control the maneuver of added background vehicles.

**Scenario 2:** Simulates the model's performance under an aggressive and sudden lead-vehicle deceleration maneuver. A safety evaluation was conducted under these safety-critical conditions, as depicted in **Fig. 11(c)**. In this test, the leading vehicle (yellow vehicle), located approximately 100 to 300 m ahead of the controlled ego vehicle in the same lane, was programmed to perform emergency braking, remain stationary for 3s, and then gradually resume its normal speed, thus simulating a sudden breakdown.

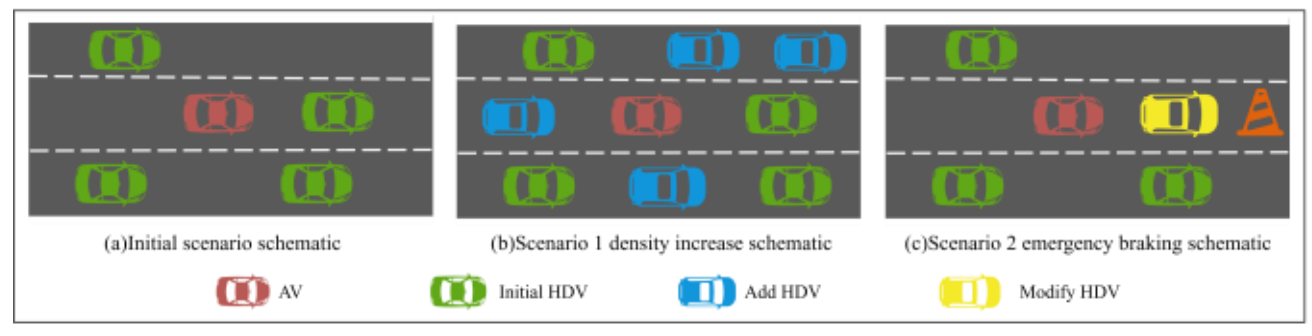

**Fig.11.** Schematics of different scenarios

The test performance regarding speed and acceleration for **Scenario 1** are shown in **Fig. 12** and **TABLE III**. **Fig. 12** reveals that the speeds of all algorithms decline with increasing traffic density, whereas the dispersion of acceleration distributions widens accordingly. In this situation, the speed distribution of the proposed KDDRL model remains higher than that of other two comparison algorithms, and the acceleration distribution remain relatively concentrated with a low deviation. **TABLE III** further quantifies the comprehensive performance of each algorithm in terms of efficiency and safety. In terms of average speed, KDDRL outperforms the other two algorithms, indicating that it can maintain relatively high driving efficiency in congested traffic conditions. In terms of average acceleration, the proposed model exhibits a higher mean value than the comparison algorithms along the lowest deviation, suggesting that its decision-making strategy actively adapt vehicle speed to complex dynamic surroundings rather than remaining passively at a low speed.

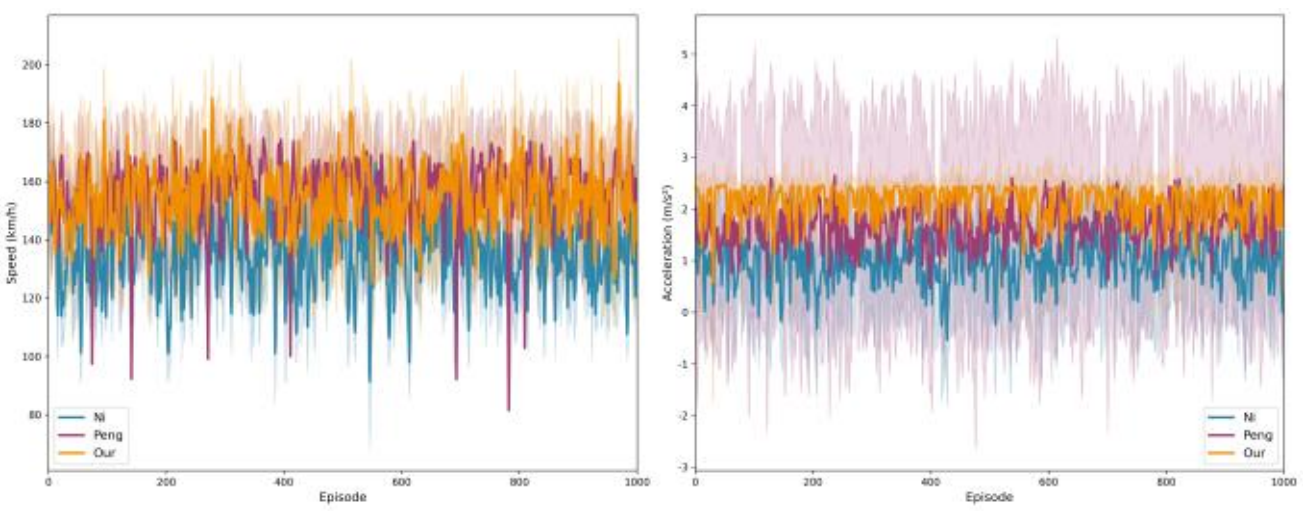

(a)Speed distribution curve (b)Acceleration distribution curve

**Fig.12.** Speed and acceleration distributions of different algorithms in the increased density scenario (Scenario 1)

TABLE III
PERFORMANCE COMPARISON OF DIFFERENT ALGORITHMS IN THE CONGESTED SCENARIO 1

| Algorithm | Average speed (km/h) | Average acceleration (m/s²) | Number of lane changes | Collision rate |
|---|---|---|---|---|
| DDQN+TD3(Ni ) | 136.3±24.1 | 0.99 ± 0.9 | 837±24 | 1.8% |
| D3QN+DDPG (Peng) | 152.5±21.4 | 1.69 ± 2.2 | 1202±27 | 1.8% |
| D3QN+TD3 (KDDRL) | **154.4±20.2** | **2.13 ± 0.5** | 1081±29 | **0.3%** |

Regarding lane-changing maneuver, Ni et al. (DDQN+TD3) executes the lowest number of lane changes (837 times), but exhibits a relatively high collision rate (1.8%). This indicates that while this algorithm tends to adopt a conservative lane-changing strategy in high-density scenarios, its safety does not improve proportionately due to imprecise judgment of lane-changing timing. Peng et al. (D3QN+DDPG) records the highest frequency of lane changes (1202 times) with a collision rate of 1.8%, revealing a relatively aggressive lane-changing strategy that improves localized speed but incurs elevated safety risks.

The proposed model achieves a more balanced tradeoff between the frequency of lane changes and the collision rate (0.3%). While ensuring lane-changing flexibility to maintain traffic efficiency, it achieves a low collision rate, supporting its potential in enhancing both safety and driving stability. In this scenario, characterized by frequent vehicle interactions and a constrained decision space, the proposed trajGAN module can predict intention aware trajectories of surrounding leading vehicles, enhancing the forward-looking nature of the state representation and enabling the ego AV to avoid potential conflicts in congested traffic. Concurrently, the knowledge-data dual-driven training paradigm incorporates lcGAN-based expert experience and MOBIL-derived physical constraints, effectively narrowing the exploration space and suppressing dangerous actions. This allows the model to significantly reduce collision risk while maintaining traffic efficiency.

In Scenario 2, as shown in **Fig. 13** and **TABLE IV**, Ni et al. (2024) records the lowest average speed (132.88 km/h) and the fewest lane changes (525 times), alongside the highest collision rate (24.8%). This indicates that its overly conservative strategy struggles to effectively balance traffic efficiency and safety when encountering sudden obstacles. Peng et al. (2022) achieves a moderate improvement in average speed (140.22 km/h) and lane-change frequency (870 times), but its collision rate remains high at 24.1%, suggesting that enhanced maneuverability alone does not translate into meaningful safety gains. In contrast, the proposed KDDRL demonstrates superior overall performance across these metrics, maintaining high traffic efficiency while ensuring safety.

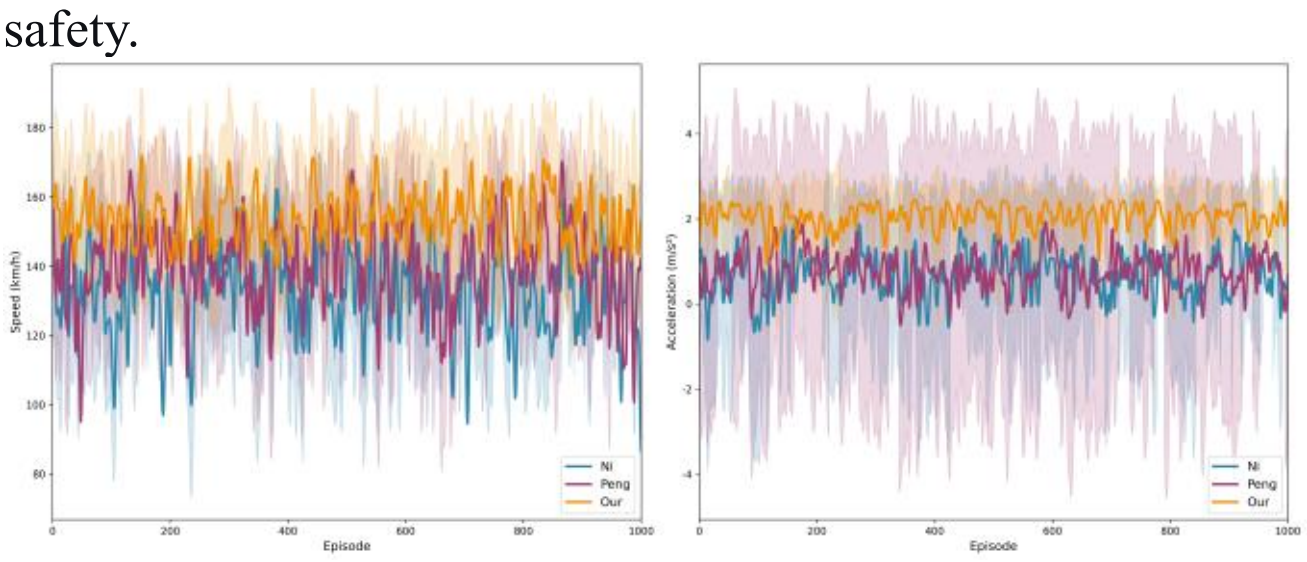

(a)Speed distribution curve (b)Acceleration distribution curve

**Fig.13.** Speed and acceleration distributions of different algorithms in the emergency braking (Scenario 2)

TABLE IV
PERFORMANCE COMPARISON OF DIFFERENT ALGORITHMS IN EMERGENCY BRAKING SCENARIO 2

| Algorithm | Average speed (km/h) | Average acceleration (m/s²) | Number of lane changes | Collision rate |
|---|---|---|---|---|
| DDQN+TD3(Ni ) | 132.9±22.1 | 0.67±2.1 | 525 ± 19 | 24.8% |
| D3QN+DDPG (Peng) | 140.2 ±23.2 | 0.78± 3.2 | 870 ± 20 | 24.1% |
| D3QN+TD3 (KDDRL) | **153.9±20.7** | **2.04 ± 0.7** | 1226 ± 29 | **1.0%** |

The lowest acceleration dispersion reflects a more sensitive and stable response capability under emergency conditions. In Scenario 2, the cumulative number of lane changes across all vehicles during 1000 episodes reaches 1,226 times, indicating that the algorithm can more actively adjust its driving strategy when facing sudden obstacles, effectively mitigating potential rear-end risks through timely lateral actions. Notably, the collision rate 1.0% is much lower than those of the other two baselines, verifying the enhanced safety performance of the proposed model under safety-critical conditions.

In this highly uncertain, safety critical scenario, the proposed KDDRL leverage a hierarchical, heterogeneous-frequency joint RL architecture to effectively decouple lane-changing and car-following decisions: D3QN generates rapid evasive decisions while TD3 facilitates promptly and smooth acceleration adjustments. By synergistically combining forward-looking, intention- aware trajectory prediction with physical safety constraints, KDDRL substantially improves the AV's emergency responsiveness under extreme traffic conditions.

To illustrate the decision-making logic and driving characteristics of KDDRL across different traffic scenarios, the spatiotemporal trajectories of the ego AV are visualized in **Fig. 14**. In the figure, the x-axis denotes the vehicle's lateral position, the y-axis represents its longitudinal position, and the z-axis corresponds to the time step.

**Fig.14(a)** depicts a high-density condition in Scenario 1 without a lane change. When the surrounding traffic density is high and insufficient space exists in adjacent lanes, the model assesses that the risk of changing lanes outweighs the potential efficiency gains and opts to remain in the current lane, following the preceding vehicle steadily. **Fig.14(b)** shows a high-density condition in Scenario 1 in which a lane change is executed. When adequate space becomes available in an adjacent lane, the leading vehicle ahead is moving slowly, and the safety constraints for lane changing are satisfied, the ego AV performs a smooth lateral shift. After completing the maneuver, it promptly returns to straight driving, improving traffic efficiency while maintaining safety. **Fig.14(c)** illustrates an emergency braking event in Scenario 2 without a lane change. Faced with the leading vehicle's sudden braking and recognizing that no suitable adjacent lane is available due to surrounding traffic, the ego AV decelerates gradually, waits for the leading vehicle to regain speed, and then follows it through the road section, demonstrating a cautious and compliant response dictated by the physical safety constraints. **Fig.14(d)** presents an emergency braking even in Scenario 2 with a lane change. When a safe lane-changing gap appears in an adjacent lane, the ego AV proactively executes a lane change to avoid the hazard. After bypassing the stopped leading vehicle, it resumes normal driving. The lane change is executed decisively, and the trajectory remains continuous and stable, achieving a dynamic trade-off between efficiency and safety.

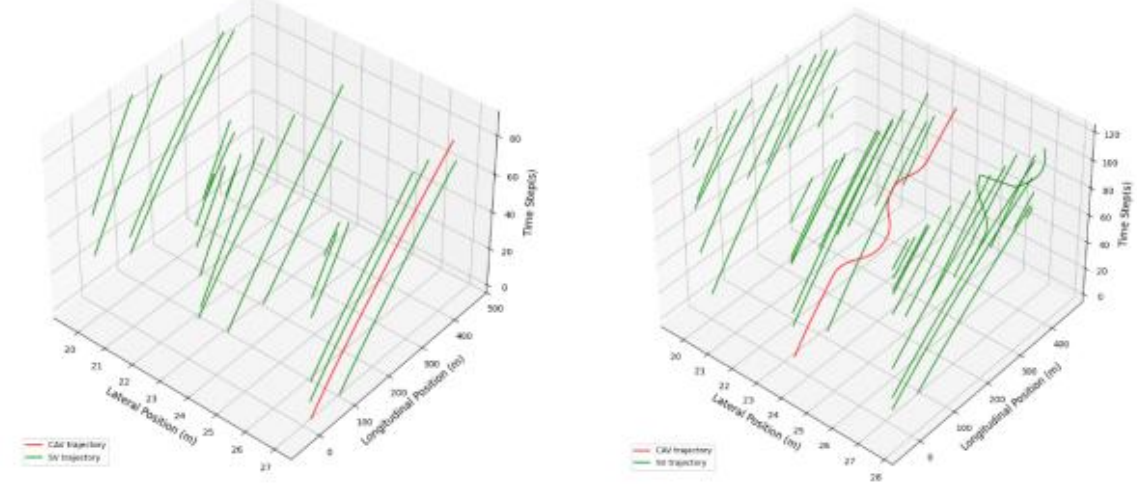

(a)Without lane change in Scenario 1 (b)Lane change in Scenario 1

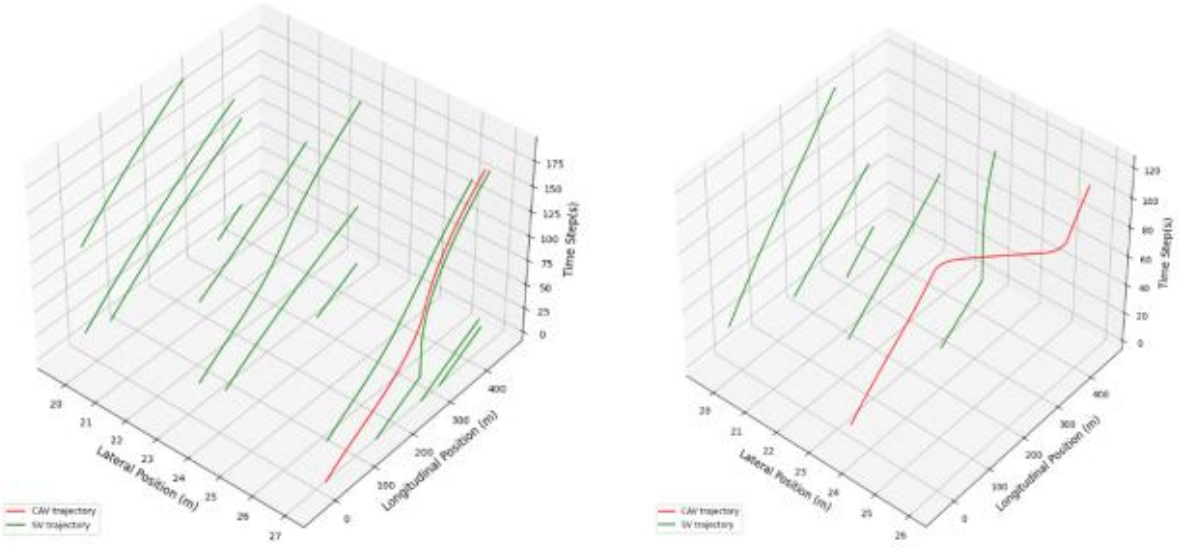

(c)Without lane change in Scenario 2 (d)Lane change in Scenario 2

**Fig.14.** Spatiotemporal trajectories of AV in different scenarios

The spatiotemporal trajectories confirm that the KDDRL framework navigates the safety-efficiency trade-off adaptively based on real-time traffic conditions. In high-density congested conditions (Scenario 1), it engages in sustained car-following and strategically delays lane changes until acceptable inter-vehicle gaps emerge. In emergency hazardous conditions (Scenario 2), it prioritizes evasive actions through rapid responses and proactive hazard avoidance. The resulting behavioral policy exhibits defensible and intuitive tactical logic while delivering enhanced safety assurance under dynamic constraints.

### *E. Ablation Experiment*

To verify the individual and synergetic effectiveness of the intention-aware trajectory generation submodule (TG) and the knowledge-guided lane-changing behavior generation submodule (LG) , ablation experiments are conducted by progressively removing or integrating the corresponding submodules within the KDDRL framework. The results are presented in **Fig. 15** and **Table V**.

**Fig. 15** reports the average training rewards in the ablation experiment. The classical rule-based model (IDM+MOBIL) yields the lowest mean and median reward values. The introduction of a pure RL algorithm (D3QN+TD3) substantially improves the mean reward, verifying the optimization capability of the RL framework for decision-making. Retaining the LG module on this basis (i.e., D3QN+TD3+LG) further enhances local state perception and interaction modeling. Ultimately, the KDDRL framework incorporating both the TG and LG modules achieves the optimal mean and median values for both lane-changing and car-following tasks, demonstrating that the TG module effectively elevates the robustness and performance ceiling of global decision-making.

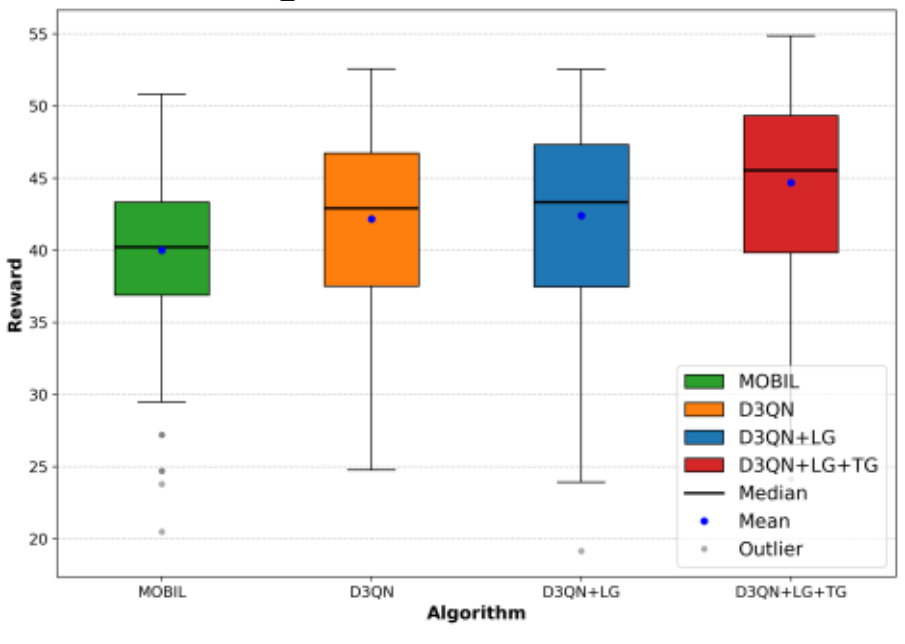

(a)Lane-changing mode

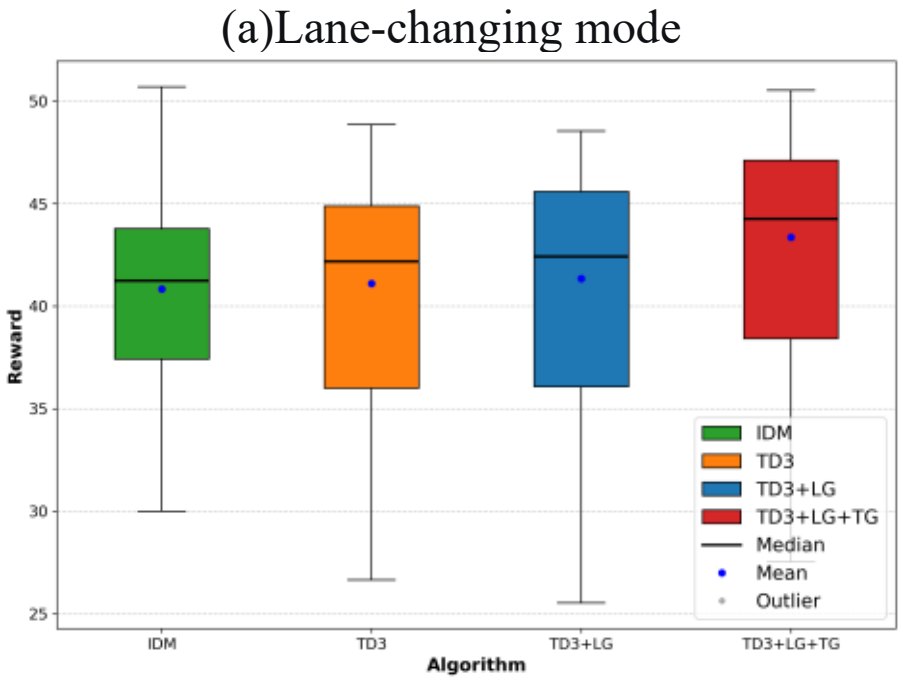

(b)Car-following model

**Fig.15.** Box plot of the average training reward across configurations in the ablation experiment

Based on the performance metrics presented in **TABLE V**, the IDM+MOBIL baseline achieves a zero collision rate but also the lowest average speed among all configurations, reflecting the conservative nature of rule-driven strategies that prioritize safety at the expense of efficiency. The pure RL model (D3QN+TD3) raises the average speed substantially but does so at the cost of a 3.4% collision rate, illustrating that purely data-driven methods can sacrifice safety in pursuit of higher efficiency.

When the LG module is added (D3QN+TD3+LG), the collision rate drops to 2.6% while the average speed increases further, demonstrating that the prior knowledge embedded in LG effectively constrains the action space and mitigates collision risk without sacrificing responsive acceleration. Integrating the TG module on top of this to form the full KDDRL framework yields the most balanced overall performance: the highest average speed among all configurations and a collision rate of 0.1% in this baseline environment, matching the safety record of the classical IDM+MOBIL baseline. This outcome confirms that the TG module, by generating diverse and realistic intention-conditioned trajectories, substantially enhances the model's ability to navigate complex traffic situations while maintaining high efficiency and collision-free operation.

Furthermore, a performance comparison was conducted with a variant of KDDRL operating at a uniform control frequency (0.1 s for both lane change and car following control). This single-frequency variant yields a lower average speed and a higher collision rate than the proposed dual-frequency KDDRL. This confirms that the hierarchical, heterogeneous-frequency control strategy effectively decouples high-level tactical choices from low-level continuous execution, providing critical stability.

In summary, the ablation results verify the individual and synergistic effectiveness of the LG and TG modules, and multi-frequency design, demonstrating that the proposed KDDRL framework achieves an optimal balance between traffic efficiency and safety.

TABLE V
EVALUATION OF ABLATION TEST PERFORMANCE

| Algorithm | Average speed (km/h) | Average acceleration ($m/s^2$) | Collision rate |
|---|---|---|---|
| IDM+MOBIL | 146.6 ± 25.5 | 1.20 ± 0.8 | 0% |
| D3QN+TD3 | 152.5 ± 30.6 | 2.42 ± 0.6 | 3.4% |
| D3QN+TD3+LG | 157.3 ± 30.4 | 2.42 ± 0.6 | 2.6% |
| D3QN+TD3+TG+LG (KDDRL) | **159.4 ± 22.5** | **2.33 ± 0.3** | **0.1%** |
| KDDRLwith same control frequency | 157.7 ± 27.2 | 1.66 ± 0.5 | 0.9% |

### *F. Training settings and results of generative adversarial networks*

#### *(1) Predicted results of trajGAN model*

For training trajGAN model, we extracted 2,325 vehicles trajectories from the HighD dataset to construct the real naturalistic trajectory dataset. All samples were randomly divided into a training set, validation set and a test set at a ratio of 8:1:1. To quantitatively evaluate the trajectory prediction accuracy of trajGAN in our scenario, this study uses Mean Absolute Error (MAE) and Root Mean Square Error (RMSE) as primary evaluation metrics. Consequently, the trajGAN model achieves a competitive trajectory prediction accuracy, yielding an MAE of 4.39 m and RMSE of 5.36 m over the evaluation horizon, as reference in Xu et al. (2025). The minimal deviation between the generated and real trajectories confirms that trajGAN can faithfully capture the motion patterns of vehicles in the real-world driving data, thereby providing a reliable forward-looking state representation for the subsequent RL-based decision-making module.

*(2) Predicted results of lcGAN model*

From vehicle trajectories recorded in the HighD dataset, we extracted lane-changing maneuvers using a 2-second sliding time window, yielding a total of 194,115 samples. These samples were randomly divided into training, validation, and test sets at a ratio of 8:1:1. **Fig.16** presents a pie chart comparing the distribution of real lane-changing maneuvers with those predicted by the lcGAN on the test set. The predicted distributions for the three maneuver types align closely with the ground truth, with lane-keeping achieving higher accuracy than lane-changing maneuvers due to its higher representation in naturalistic data. The overall correct prediction accuracy reaches 98.13%, verifying the model's generalization capability in terms of both behavioral distribution consistency and prediction accuracy.

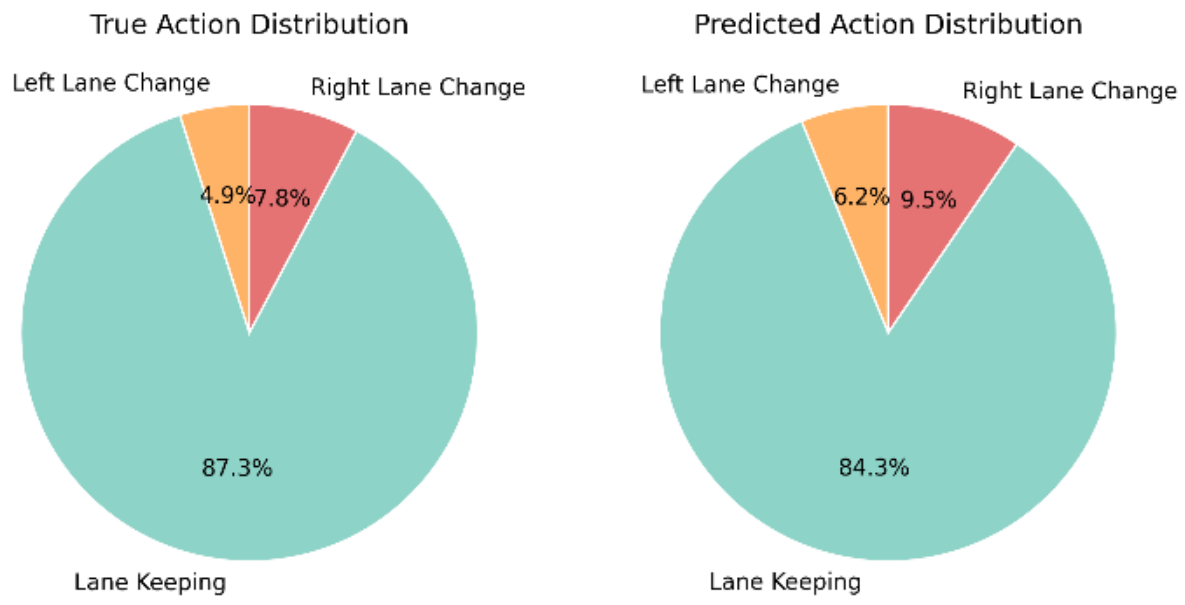

**Fig.16.** Distribution of model lane-changing actions during testing

**Fig.17** presents the confusion matrix for lane-changing maneuvers predicted by the lcGAN. The diagonal elements indicate the proportion of correctly predicted actions: 97.2% for left lane changes, 96.0% for lane keeping, and 95.7% for right lane changes. Misclassifications are primarily symmetric errors between left and right lane changes, with minimal confusion between lane-changing and lane-keeping maneuvers. The overall recall is 96.02%, demonstrating the model's capability to reliably identify driving intentions and distinguish lane-changing from lane-keeping. These results confirm the effectiveness and robustness of the lcGAN for lane-changing decision prediction.

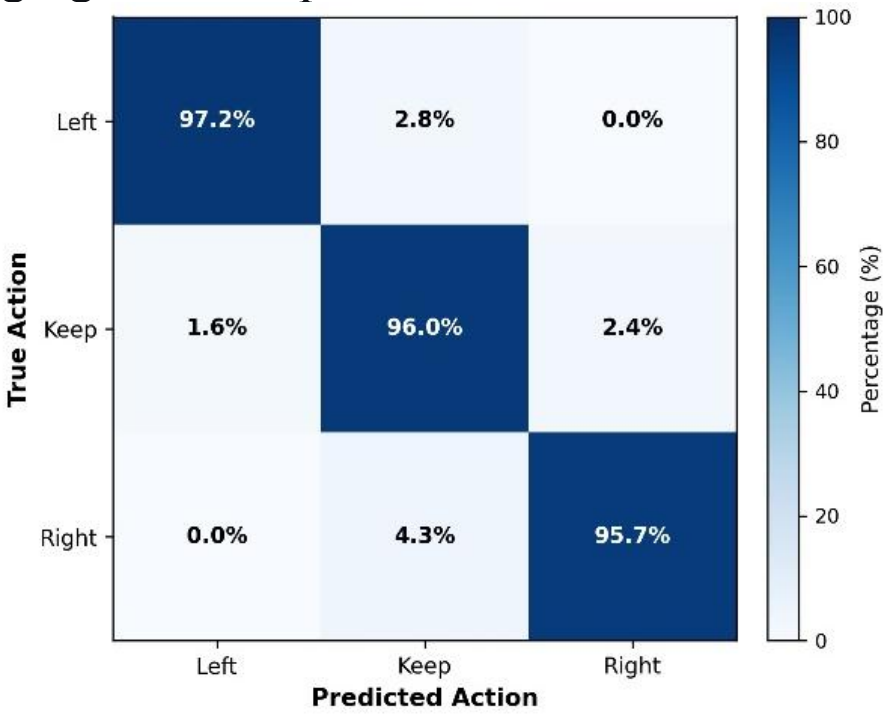

**Fig.17.** Action confusion recall matrix

The accurate prediction outputs from trajGAN and lcGAN contribute substantially to the effectiveness of overall decision-making framework. First, the precise intention-aware trajectory predictions produced by trajGAN expand the reliable and anticipatory state representations to RL agent. This anticipatory awareness mitigate the conservative efficiencies inherent in safety-critical scenarios, enabling the RL agents to foresee potential conflicts and execute proactive maneuvers rather than reacting passively to immediate observations.

Second, the high-fidelity lane-change intention predictions from lcGAN supply trustworthy, data-driven priors. When synergistically combined with physics-based MOBIL constraints, these priors effectively narrow the agent's exploration space, accelerate policy convergence, and endow the learned policy with both empirical rationality and physical feasibility.

Collectively, the superior predictive capabilities of the two generative models form the cornerstone of the KDDRL framework, allowing the hierarchical joint architecture to seamlessly integrate uncertainty-aware perception, knowledge-guided learning, and heterogeneous action coordination. This reliable perception of future vehicle intentions enables the framework to achieve a balanced trade-off between traffic efficiency and safety, as demonstrated in the evaluations.

## V. CONCLUSION

Safe and efficient decision-making for a single ego automated vehicle in mixed traffic flows, where it must interact with multiple human-driven vehicles exhibiting stochastic and intention-dependent maneuvers, remains a fundamental challenge for autonomous driving. This paper proposes a Knowledge-Data Dual-Driven Reinforcement Learning (KDDRL) framework to address this challenge through three synergistic innovations.

First, an improved intention-aware trajectory generation model based on a conditional generative adversarial network (trajGAN) encodes interactive vehicle intentions and synthesizes multiple physically plausible future trajectories. This design converts passive perception into proactive, intention-aware predictive states, overcoming the limitations of deterministic physics models in capturing diverse human driving behaviors under identical traffic conditions.

Second, a knowledge-data dual-driven paradigm operates on these predictive states by fusing data-driven probabilistic insights with interpretable physical boundaries as safety-efficiency inductive biases. This hybrid structure stabilizes policy optimization across non-stationary traffic regimes and accelerates convergence in rare safety-critical events without sacrificing adaptability to human driving heterogeneity.

Third, a unified predictive-knowledge state basis compresses intention-aware trajectories and physics-based constraints into compact shared embeddings. This enables asynchronous multi-timescale optimization, wherein discrete lane-changing decisions (D3QN) and continuous car-following control (TD3) operate at their natural temporal resolutions while preserving interdependent state information, thus resolving the policy instability inherent in hybrid action spaces.

Comprehensive experiments on simulation environments calibrated with real-world HighD trajectory data validate the efficacy of the proposed framework: the trajGAN achieves accurate trajectory prediction, while the lcGAN attains 98.13% classification accuracy in lane-change intention recognition. Across normal, congested, and emergency-braking scenarios, KDDRL consistently outperforms state-of-the-art baselines in safety, efficiency, and comfort, reducing collision rates while maintaining higher average speed. Ablation studies further confirm the synergistic necessity of the trajectory generation and LC behavior generation submodules, as well as multi-frequency control design.

Future work will explore online adaptive updating mechanisms for the generative modules to address distribution shifts in unobserved environments, incorporate green driving metrics such as energy consumption into a multi-objective optimization framework, and extend the current single-ego-AV paradigm to multi-vehicle cooperative interaction control through V2X communication.

## APPENDIX

TABLE A.1
EXPERIMENTAL PARAMETERS FOR SUMO SIMULATING MIX TRAFFIC SCENARIOS

| Parameters | Value |
|---|---|
| **Simulator** | |
| Number of lanes | 3 |
| Lane length | 420 $m$ |
| Number of vehicles | 2325 $veh/h$ |
| Minimum speed | 72 $km/h$ |
| Maximum speed | 180 $km/h$ |
| Lane-changing duration | 2 $s$ |
| Simulation time step | 0.1 $s$ |
| **Model-IDM** | |
| Desired speed $v_0$ | 130 $km/h$ |
| Acceleration exponent $\delta$ | 4 |
| Desired minimum following gap $s_0$ | 5 $m$ |
| Maximum acceleration $a_{max}$ | 5 $m/s^2$ |
| Comfortable deceleration $b$ | 4 $m/s^2$ |
| Desired time headway T | 1.5 $s$ |
| **Model-MOBIL** | |
| Lane-changing threshold $\Delta a_{th}$ | 0.3 |
| Preset safe deceleration $b_{safe}$ | 4 $m/s^2$ |
| Politeness coefficient $p$ | 0.1 |

**Algorithm 1** KDDRL Framework

**1 Initialize for KDDRL:**
2 Pre-train trajGAN
3 Pre-train lcGAN
4 **Initialize D3QN network:** $Q(s, a \mid \theta)$ and target $Q(s, a \mid \theta')$
5 **Initialize TD3 networks:**
6 actor $\mu(s \mid \phi)$ with $\phi$, target actor $\mu(s \mid \phi')$ with t $\phi'$
7 critics $Q_1(s, a \mid \varphi_1)$,$Q_2(s, a \mid \varphi_2)$ and targets $Q_1'$, $Q_2'$with $\varphi_1'$, $\varphi_2'$
8 Initialize empty replay buffer ***H*** for D3QN
9 Initialize empty replay buffer ***L*** for TD3
10 Set decision at different time scales: $\Delta t_{D3QN}$= 2s, $\Delta t_{TD3}$= 0.1s, interval = $\Delta t_{D3QN}$ / $\Delta t_{TD3}$ = 20
**11 for** *episode = 1, 2, ..., M* **do**
12 Initialize initial state $s_0$ from Environment
13 **for** *t = 1, 2, ..., T* **do**
14 while not terminal and no collision do
15 Compute IDM initial trajectory $TC^T$
16 Sample noise $Z^T \sim \mathcal{N}(0,1)$, generate $TG^T$
17 Extract $TG^{t10}$,$TG^{t20}$, compute$(TG^{t10}-\boldsymbol{y}_{AV}^{t})$ $(TG^{t20}-\boldsymbol{y}_{AV}^{t})$
18 Construct condition $C^t$,Generate lc probability $\boldsymbol{P}_{lcGAN}$
19 Check safety and compute $\boldsymbol{U_d}$ according to (20)-(21)
20 Obtain $\boldsymbol{P}_{MOBIL}$ according to (22)
21 Fuse prior probability $\boldsymbol{P}_{lc}$
22 **if** *t* mod interval == 0 **then**
23 Obtain $s_{D3QN}^t$ according to (24)
24 Select $a_{D3QN}$ = argmax $Q(s_{D3QN}^t, a \mid \theta)$ with ε-greedy
25 Store transition [$s_{D3QN}^t$, $a_{D3QN}$,$r_{\text{D3QN}}$, $s_{D3QN}^{t+1}$] in ***H***
26 cumulative $r_{\text{D3QN}}$ = 0
27 Sample minibatch ***B*** from ***H***
28 Using gradient descent to update the D3QN $Q(s, a \mid \theta)$
29 Update D3QN target network $\theta' \leftarrow (1-\tau)\theta' + \tau\theta$
30 **end**
31 Obtain state $s_{TD3}^t$ according to (30):
32 Select $a_{TD3} = \mu(s_{TD3}^t \mid \phi) + clip(\varepsilon, -c, c)$
33 Execute $a_{D3QN}$ and $a_{TD3}$ in Env
34 Compute reward $r_{D3QN}$,$r_{TD3}$ according to (26)(32)
35 cumulative $r_{\text{D3QN}}$ += $r_{\text{D3QN}}$
36 Observe next state $s^{t+1}$
37 Store transition [$s_{TD3}^t$, $a_{TD3}$,$r_{TD3}$,$s_{TD3}^{t+1}$] in ***L***
38 Sample minibatch ***B*** from ***L***
39 Update critics network with $\varphi_1$, $\varphi_2$ according to (6)-(7)
40 Delayed updates of actor and targets:
41 Update $\phi$ via deterministic policy gradient according to (8)
42 Update target networks:
43 $\varphi_k' \leftarrow (1-\tau)\varphi_k' + \tau\varphi_k$,
$\phi' \leftarrow (1-\tau)\phi' + \tau\phi$
44 **end for**
45 **end for**

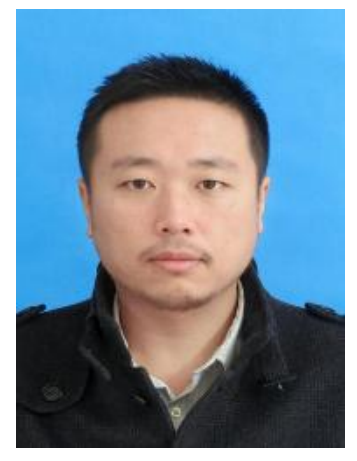

**Jie Fang** received his Ph.D degrees in the Civil and Environmental Engineering department from University of Wisconsin-Madison, U.S. He is currently a Professor in the College of Civil Engineering, Fuzhou University. His research interests include proactive traffic control and deep learning, machine learning.

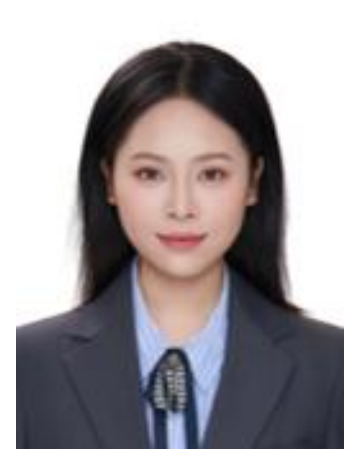

**Wei Zheng** is currently pursuing the M.S. degree in transportation engineering with the College of Civil Engineering, Fuzhou University, Fuzhou, China. Her research interests include autonomous vehicle control, reinforcement learning and deep learning.

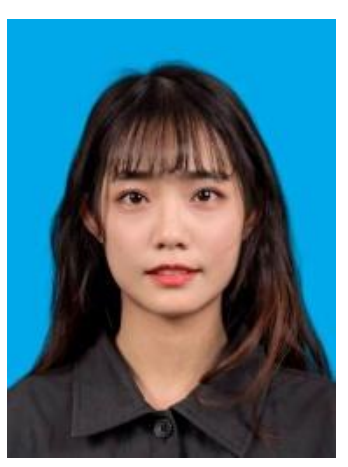

**Mengyun Xu** received her Ph.D degrees in the Civil and Environmental Engineering department from University of Wuhan technology. She is an associate research fellow in the college of civil engineering, Fuzhou University as well as in the civil and environmental engineering, National University of Singapore. Her research interest is in intelligent transportation systems.

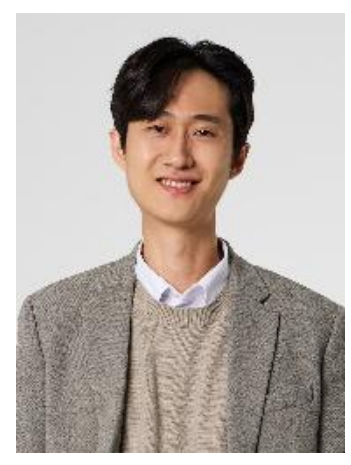

**Eui-Jin Kim** received the Ph.D. degree in civil and environmental engineering from Seoul National University, Seoul, Korea, in 2021. He is an Assistant Professor at the Department of Transportation Systems Engineering, Ajou University, Suwon, Republic of Korea. His research interests include artificial intelligence, intelligent transportation system, and travel behavior.